\documentclass[nohyperref]{article}

\usepackage{microtype}
\usepackage{graphicx}
\usepackage{subfigure}
\usepackage{booktabs} 

\usepackage[colorlinks=true,linkcolor=Blue9,citecolor=Blue9]{hyperref}
\usepackage{url}
\usepackage{graphicx}
\usepackage{amsfonts, amssymb, amsthm, amsmath}
\usepackage{mathtools}
\usepackage{wrapfig}
\usepackage{overpic}
\usepackage{subfigure}
\usepackage{caption}
\usepackage{booktabs}    
\usepackage{multirow}

\usepackage{algorithm}
\usepackage{algpseudocode}

\usepackage{hyperref}

\usepackage[accepted]{icml2022}

\usepackage{amsmath}
\usepackage{amssymb}
\usepackage{mathtools}
\usepackage{amsthm}

\usepackage{listings}
\usepackage{xcolor}

\definecolor{codegreen}{rgb}{0,0.6,0}
\definecolor{codegray}{rgb}{0.5,0.5,0.5}
\definecolor{codepurple}{rgb}{0.58,0,0.82}
\definecolor{backcolour}{rgb}{0.95,0.95,0.92}

\definecolor{mygreen}{rgb}{0,0.6,0}
\definecolor{mygray}{rgb}{0.5,0.5,0.5}
\definecolor{mymauve}{rgb}{0.58,0,0.82}
\lstdefinestyle{mystyle}{
    backgroundcolor=\color{backcolour},   
    commentstyle=\color{codegreen},
    keywordstyle=\color{magenta},
    numberstyle=\tiny\color{codegray},
    stringstyle=\color{codepurple},
    basicstyle=\ttfamily\footnotesize,
    breakatwhitespace=false,         
    breaklines=true,                 
    captionpos=b,                    
    keepspaces=true,                 
    numbers=left,                    
    numbersep=5pt,                  
    showspaces=false,                
    showstringspaces=false,
    showtabs=false,                  
    tabsize=2
}

\usepackage{tikz}
\usetikzlibrary{backgrounds}
\usetikzlibrary{arrows,shapes}
\usetikzlibrary{tikzmark}
\usetikzlibrary{calc}

\usepackage[capitalize,noabbrev]{cleveref}

\theoremstyle{plain}

\theoremstyle{definition}

\theoremstyle{remark}

\usepackage[textsize=tiny]{todonotes}

\usepackage{stackengine}

\newcommand{\alert}[1]{\textbf{}}

\newcommand{\BEAS}{\begin{eqnarray*}}
\newcommand{\EEAS}{\end{eqnarray*}}
\newcommand{\BEA}{\begin{eqnarray}}
\newcommand{\EEA}{\end{eqnarray}}
\newcommand{\BEQ}{\begin{equation}}
\newcommand{\EEQ}{\end{equation}}
\newcommand{\BIT}{\begin{itemize}}
\newcommand{\EIT}{\end{itemize}}
\newcommand{\BNUM}{\begin{enumerate}}
\newcommand{\ENUM}{\end{enumerate}}
\newcommand{\BEL}[1]{\begin{equation}\label{#1}}
\newcommand{\EEL}{\end{equation}}

\newcommand{\namedef}{Contrastive Intrinsic Control (CIC) }
\newcommand{\name}{CIC }

\newcommand{\state}{s}

\newcommand{\skill}{z}

\newcommand{\nextstate}{s'}

\newcommand{\action}{a}

\newcommand{\reward}{r}

\newcommand{\BA}{\begin{array}}
\newcommand{\EA}{\end{array}}

\DeclareMathOperator*{\expec}{\mathbb{E}}

\icmltitlerunning{CIC: Contrastive Intrinsic Control for Unsupervised Skill Discovery}

\begin{document}

\twocolumn[
\icmltitle{CIC: Contrastive Intrinsic Control for Unsupervised Skill Discovery}




\begin{icmlauthorlist}
\icmlauthor{Michael Laskin}{ucb}
\icmlauthor{Hao Liu}{ucb}
\icmlauthor{Xue Bin Peng}{ucb}
\icmlauthor{Denis Yarats}{nyu,fair}
\icmlauthor{Aravind Rajeswaran}{fair}
\icmlauthor{Pieter Abbeel}{ucb,cov}

\end{icmlauthorlist}

\icmlaffiliation{ucb}{UC Berkeley}
\icmlaffiliation{nyu}{NYU}
\icmlaffiliation{fair}{MetaAI}
\icmlaffiliation{cov}{Covariant}

\icmlcorrespondingauthor{Michael Laskin}{mlaskin@berkeley.edu}

\icmlkeywords{Machine Learning, ICML}

\vskip 0.3in
]



\printAffiliationsAndNotice{} 

\begin{abstract}

We introduce Contrastive Intrinsic Control (CIC), an algorithm for unsupervised skill discovery that maximizes the mutual information between state-transitions and latent skill vectors. CIC utilizes contrastive learning between state-transitions and skills to learn behavior embeddings and maximizes the entropy of these embeddings as an intrinsic reward to encourage behavioral diversity. We evaluate our algorithm on the Unsupervised Reinforcement Learning Benchmark, which consists of a long reward-free pre-training phase followed by a short adaptation phase to downstream tasks with extrinsic rewards. CIC substantially improves over prior methods in terms of adaptation efficiency, outperforming prior unsupervised skill discovery methods by $1.79\times$ and the next leading overall exploration algorithm by $1.18\times$.\footnote{Project website and code: \url{https://sites.google.com/view/cicrl/}}
\end{abstract}

\section{Introduction}

\begin{figure}[t]

\begin{center}
\centerline{\includegraphics[width=.75\columnwidth]{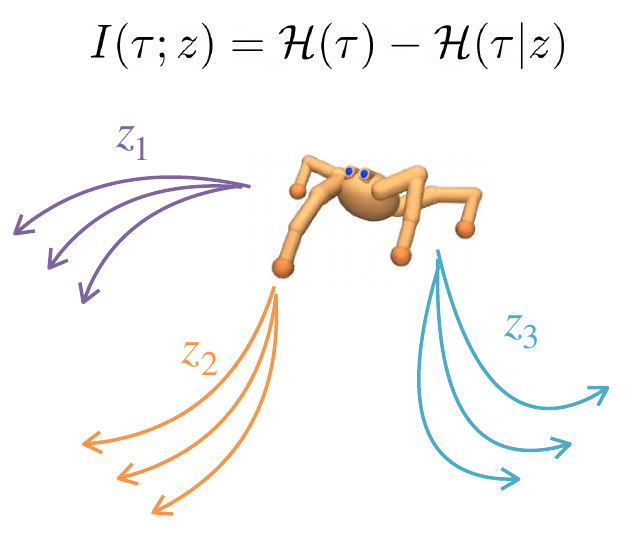}}
\end{center}
\caption{This work deals with unsupervised skill discovery through mutual information maximization. We introduce \namedef -- a new unsupervised RL algorithm that explores and adapts more efficiently than prior methods.}
\label{fig:teaser}
\end{figure}

Deep Reinforcement Learning (RL) is a powerful approach toward solving complex control tasks in the presence of extrinsic rewards. Successful applications include playing video games from pixels ~\citep{mnih2015human}, mastering the game of Go~\citep{alphago,alphazero}, robotic locomotion~\citep{schulman2015high,ppo,Peng2018DeepMimic} and dexterous manipulation~\citep{Rajeswaran-RSS-18, OpenAIHand, OpenAI2019SolvingRC} policies. While effective, the above advances produced agents that are unable to generalize to new downstream tasks beyond the one they were trained to solve. Humans and animals on the other hand are able to acquire skills with minimal supervision and apply them to solve a variety of downstream tasks. In this work, we seek to train agents that acquire skills without supervision with generalization capabilities by efficiently adapting these skills to downstream tasks.

Over the last few years, unsupervised RL has emerged as a promising framework for developing RL agents that can generalize to new tasks. In the unsupervised RL setting, agents are first pre-trained with self-supervised intrinsic rewards and then finetuned to downstream tasks with extrinsic rewards. Unsupervised RL algorithms broadly fall into three categories - knowledge-based, data-based, and competence-based methods\footnote{These categories for exploration algorithms were introduced by~\citet{srinivas_abbeel_2021_icml_tutorial} and inspired by~\citet{oudeyer2007intrinsic}.}. Knowledge-based methods maximize the error or uncertainty of a predictive model~\citep{pathak2017curiosity,Pathak19disagreement,burda2018exploration}. Data-based methods maximize the entropy of the agent's visitation~\citep{liu2021unsupervised, yarats21protorl}. Competence-based methods learn skills that generate diverse behaviors~\citep{EysenbachGIL19diayn, GregorRW17vic}. This work falls into the latter category of competence-based exploration methods.

\begin{figure*} [t] \centering
\includegraphics[width=.9\textwidth]{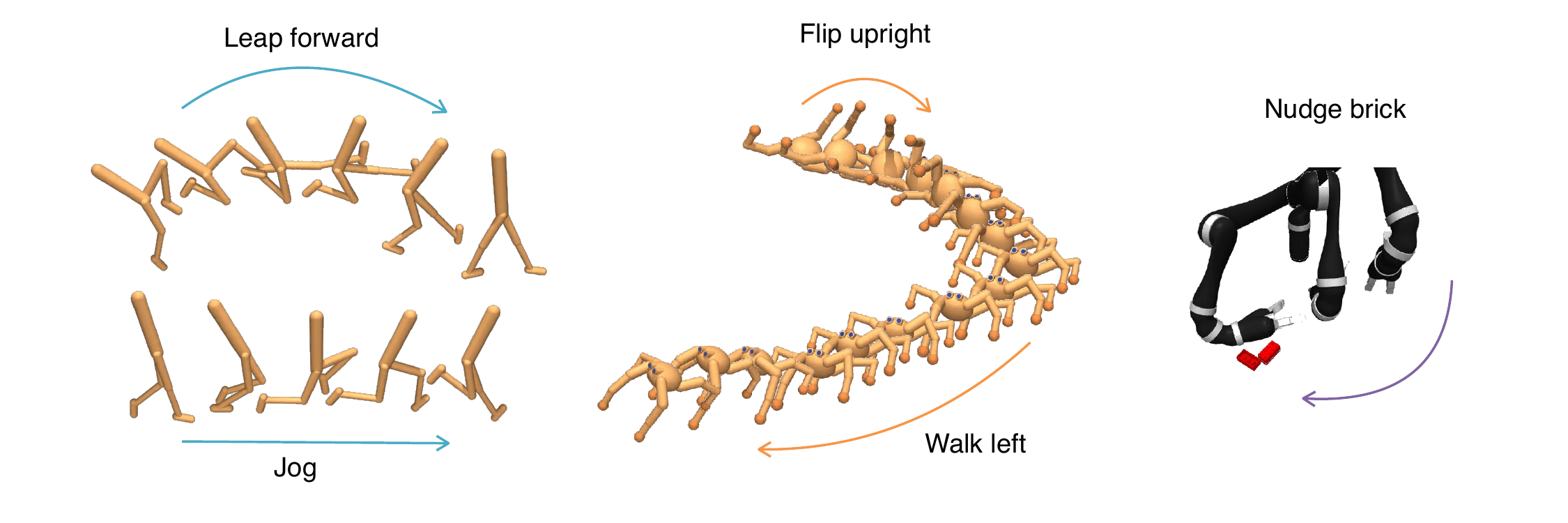}
\caption{\small Qualitative visualizations of unsupervised skills discovered in Walker, Quadruped, and Jaco arm environments. The Walker learns to balance and move, the Quadruped learns to flip upright and walk, and the 6 DOF robotic arm learns how to move without locking. Unlike prior competence-based methods for continuous control which evaluate on OpenAI Gym (e.g.~\citet{EysenbachGIL19diayn}), which reset the environment when the agent loses balance, \name is able to learn skills in fixed episode length environments which are much harder to explore (see Appendix~\ref{app:gymvsdmc}). }
\label{fig:qual_skills}
\end{figure*}

Unlike knowledge-based and data-based algorithms, competence-based algorithms simultaneously address both the exploration challenge as well as distilling the generated experience in the form of reusable skills. This makes them particularly appealing, since the resulting skill-based policies (or skills themselves) can be finetuned to efficiently solve downstream tasks. While there are many self-supervised objectives that can be utilized, our work falls into a family of methods that learns skills by maximizing the mutual information between visited states and latent skill vectors. Many earlier works have investigated optimizing such objectives~\citep{EysenbachGIL19diayn, GregorRW17vic, kwon2021variational, SharmaGLKH20dads}. 
However, competence-based methods have been empirically challenging to train and have under-performed when compared to knowledge and data-based methods~\citep{laskin_yarats_2021_urlb}.

In this work, we take a closer look at the challenges of pre-training agents with competence-based algorithms. We introduce \namedef -- an exploration algorithm that uses a new estimator for the mutual information objective. \name combines particle estimation for state entropy~\citep{singh03entropy,liu2021unsupervised} and noise contrastive estimation~\citep{gutmann2019nce} for the conditional entropy which enables it to both generate diverse behaviors {\it (exploration)} and discriminate high-dimensional continuous skills {\it (exploitation)}. To the best of our knowledge, \name is the first exploration algorithm to utilize noise contrastive estimation to discriminate between state transitions and latent skill vectors. Empirically, we show that \name adapts to downstream tasks  more efficiently than prior exploration approaches on the Unsupervised Reinforcement Learning Benchmark (URLB). \name achieves $79\%$ higher returns on downstream tasks than prior competence-based  algorithms and $18\%$ higher returns than the next-best exploration algorithm overall.

\section{Background and Notation}

{\it \bf Markov Decision Process:} We operate under the assumption that our system is described by a Markov Decision Process (MDP)~\citep{sutton2018reinforcement}. An MDP consiss of the tuple $(\mathcal S, \mathcal A, \mathcal P, r, \gamma)$ which has states $\state \in \mathcal S$, actions $\action \in \mathcal A$, transition dynamics $p(\nextstate|\state,\action) \sim \mathcal P$, a reward function $r$, and a discount factor $\gamma$. In an MDP, at each timestep $t$, an agent observes the current state $\state$, selects an action from a policy $\action \sim \pi (\cdot | \state)$, and then observes the reward and next state once it acts in the environment: 
$r, \nextstate \sim \text{env.step}(\action)$. Note that usually $r$ refers to an extrinsic reward. However, in this work we will first be pre-training an agent with intrinsic rewards $r^{\text{int}}$ and finetuning on extrinsic rewards $r^{\text{ext}}$.

\begin{figure*} [t!] \centering
\includegraphics[width=.8\textwidth]{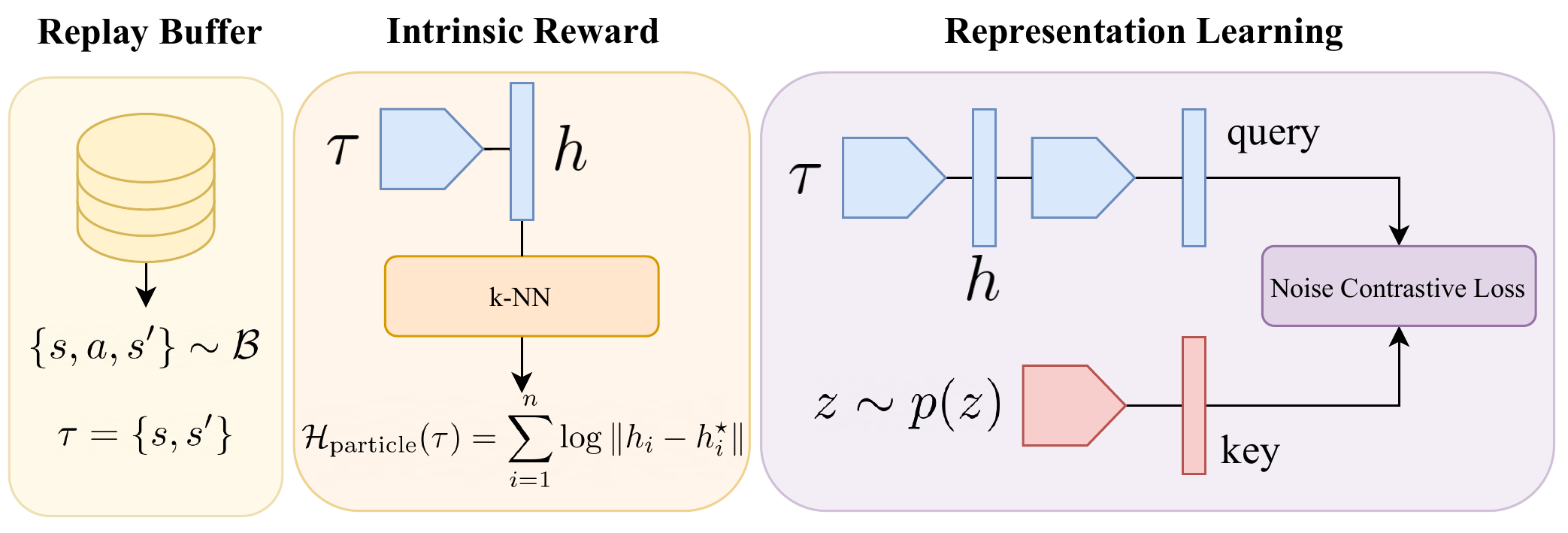}
\caption{\small Architecture illustrating the practical implementation of \name. During a gradient update step, random $\tau = (s,s')$ tuples are sampled from the replay buffer, then a particle estimator is used to compute the entropy and a noise contrastive loss to compute the conditional entropy. The contrastive loss is backpropagated through the entire architecture. The entropy and contrastive terms are then scaled and added to form the intrinsic reward. The RL agent is optimized with a DDPG~\cite{lillicrap15ddpg}. 
}
\label{fig:arch}
\end{figure*}

For convenience we also introduce the variable $\tau(s)$ which refers to any function of the states $s$. For instance $\tau$ can be a single state, a pair of states, or a sequence depending on the algorithm. Our method uses $\tau = (s, s')$ to encourage diverse state transitions while other methods have different specifications for $\tau$. Importantly, $\tau$ does not denote a state-action trajectory, but is rather shorthand for any function of the states encountered by the agent. In addition to the standard MDP notation, we will also be learning skills $\skill \in \mathcal Z$ and our policy will be skill-conditioned $\action \sim \pi (\cdot | \state, \skill)$.

{\it \bf Unsupervised Skill Discovery through Mutual Information Maximization: } Most competence-based approaches to exploration maximize the mutual information between states and skills. Our work and a large body of prior research~\citep{EysenbachGIL19diayn, SharmaGLKH20dads, GregorRW17vic, achiam2018valor, lee2019smm, liu21aps} aims to maximize a mutual information objective with the following general form:
\begin{align}
    \label{eq:mutualinfo}
    I(\tau; \skill) = \mathcal H(\skill) -  \mathcal H(\skill | \tau) = \mathcal H(\tau) -  \mathcal H(\tau | \skill)
\end{align}

Competence-based algorithms use different choices for $\tau$ and can condition on additional information such as actions or starting states. For a full summary of competence-based algorithms and their objectives see Table~\ref{table:skill_algos} in Appendix~\ref{appendix:prior_skills}.

{\it \bf Lower Bound Estimates of Mutual Information: } The mutual information $I(\state;\skill)$ is intractable to compute directly. Since we wish to maximize $I(\state;\skill)$, we can approximate this objective by instead maximizing a lower bound estimate. Most known mutual information maximization algorithms use the variational lower bound introduced in~\citet{Barber2003TheIA}:

\vspace*{-0.15in}
\begin{align}
I(\tau; \skill) = \mathcal{H}(\skill) - \mathcal{H}(\skill|\tau)  \geq \mathcal{H}(\skill) + \expec [\log q(\skill|\tau)] 
\label{eq:bound}
\end{align}
\vspace*{-0.15in}

The variational lower bound can be applied to both decompositions of the mutual information. The design decisions of a competence-based algorithm therefore come down to (i) which decomposition of $I(\tau;z)$ to use, (ii) whether to use discrete or continuous skills, (iii) how to estimate $H(\skill)$ or $H(\tau)$, and finally (iv) how to estimate $H(\skill |\tau)$ or $H(\tau|\skill)$.  

\section{Motivation}
\label{sec:problem}

Results from the recent Unsupervised Reinforcement Learning Benchmark (URLB)~\cite{laskin_yarats_2021_urlb} show that competence-based approaches underperform relative to knowledge-based and data-based baselines on DeepMind Control (DMC). We argue that the underlying issue with current competence-based algorithms when deployed on harder exploration environments like DMC has to do with the currently used estimators for $I(\tau;z)$ rather than the objective itself. To produce structured skills that lead to diverse behaviors, $I(\tau; z)$ estimators  must (i) explicitly encourage diverse behaviors and (ii) have the capacity to discriminate between high-dimensional continuous skills. Current approaches do not satisfy both criteria.

{\it Competence-base algorithms do not ensure diverse behaviors:} Most of the best known competence-based approaches~\citep{EysenbachGIL19diayn,GregorRW17vic,achiam2018valor, lee2019smm}, optimize the first decomposition of the mutual information $\mathcal{H}(\skill) - \mathcal{H}(\skill|\tau)$. The issue with this decomposition is that while it ensures diversity of skill vectors it does not ensure diverse behavior from the policy, meaning $\max \mathcal H (z)$ does not imply $\max \mathcal H(\tau)$. Of course, if $H(z) - \mathcal H(z|\tau)$ is maximized and the skill dimension is sufficiently large, then $\mathcal H(\tau)$ will also be maximized implicitly. Yet in practice, to learn an accurate discriminator $q(z|\tau)$, the above methods assume skill spaces that are much smaller than the state space (see Table~\ref{table:skill_algos}), and thus behavioral diversity may not be guaranteed. In contrast, the decomposition $I(\tau;z) = \mathcal H(\tau) - \mathcal H(\tau|z)$ ensures diverse behaviors through the entropy term $\mathcal H(\tau)$. Methods that utilize this decomposition include~\citet{liu21aps, SharmaGLKH20dads}.

\begin{figure*} [t] \centering
\includegraphics[width=\textwidth]{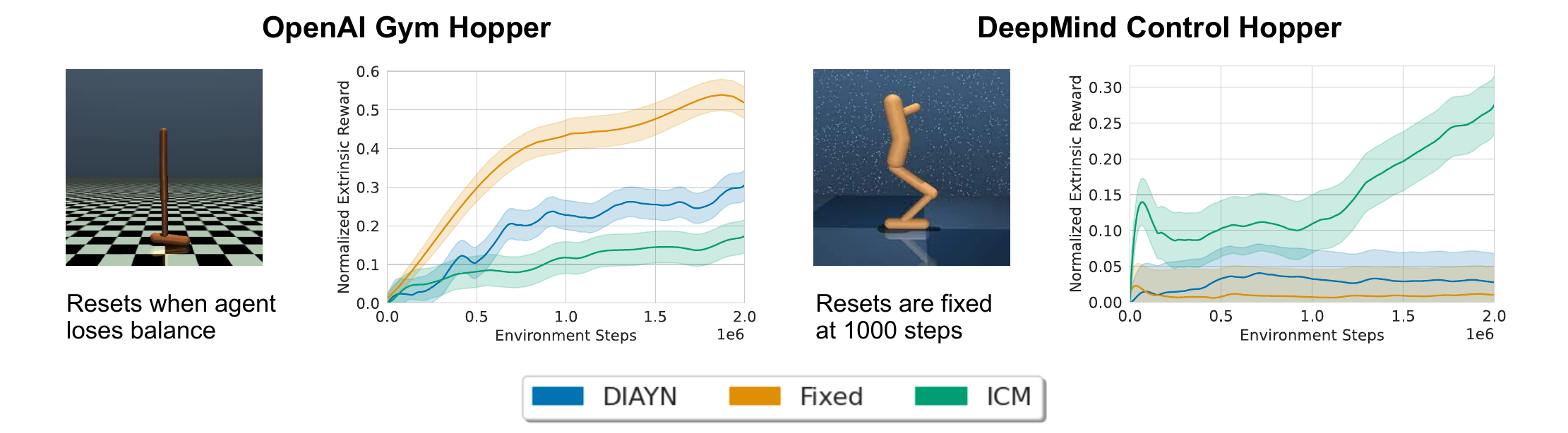}
\caption{\small To empirically demonstrate  issues inherent to competence-based exploration methods, we run DIAYN~\citep{EysenbachGIL19diayn} and compare it to ICM~\citep{pathak2017curiosity} and a {\it Fixed} baseline where the agent receives an intrinsic reward of 1.0 for each timestep and no extrinsic reward on both OpenAI Gym {\it (episode resets when agent loses balance)} and DeepMind Control (DMC) {\it (episode is fixed for 1k steps)} Hopper environments. Since Gym and DMC rewards are on different scales, we normalize rewards based on the maximum reward achieved by any algorithm (~1k for Gym, ~3 for DMC). While DIAYN is able to achieve higher extrinsic rewards than ICM on Gym, the Fixed intrinsic reward baseline performs best. However, on DMC the Fixed and DIAYN agents achieve near-zero reward while ICM does not. This is consistent with findings of prior work that DIAYN is able to learn diverse behaviors in Gym~\citep{EysenbachGIL19diayn} as well as the observation that DIAYN performs poorly on DMC environments~\citep{laskin_yarats_2021_urlb}}
\label{fig:gymvsdmcskill}
\end{figure*}

{\it Why it is important to utilize high-dimensional skills:} Once a policy is capable of generating diverse behaviors, it is important that the discriminator can distill these behaviors into distinct skills. If the set of behaviors outnumbers the set of skills, this will result in degenerate skills -- when one skill maps to multiple different behaviors. It is therefore important that the discriminator can accommodate continuous skills of sufficiently high dimension. Empirically, the discriminators used in prior work utilize only low-dimensional continuous skill vectors.  DIAYN~\citep{EysenbachGIL19diayn} utilized 16 dimensional skills, DADS~\citep{SharmaGLKH20dads} utilizes continuous skills of dimension $2-5$, while APS~\citep{liu21aps}, an algorithm that utilizes successor features~\citep{barreto2016successor, hansen20visr} for the discriminator, is only capable of learning continuous skills with dimension $10$. We show how small skill spaces can lead to ineffective exploration in a simple gridworld setting in Appendix~\ref{appendix:gridworld} and  evidence that skill dimension affects performance in Fig.~\ref{fig:skill_abs}.

{\it On the importance of benchmarks for evaluation:} While prior competence-based approaches such as DIAYN~\citep{EysenbachGIL19diayn} were evaluated on OpenAI Gym~\citep{brockman2016openai}, Gym environment episodes terminate when the agent loses balance thereby leaking some aspects of extrinsic signal to the exploration agent. On the other hand, DMC episodes have fixed length. We show in Fig~\ref{fig:gymvsdmcskill} that this small difference in environments results in large performance differences. Specifically, we find that DIAYN is able to learn diverse skills in Gym but not in DMC, which is consistent with both observations from DIAYN and URLB papers. Due to fixed episode lengths, DMC tasks are harder for reward-free exploration since agents must learn to balance without supervision.

\section{Method}

\begin{figure*} [t] \centering
\includegraphics[width=\textwidth]{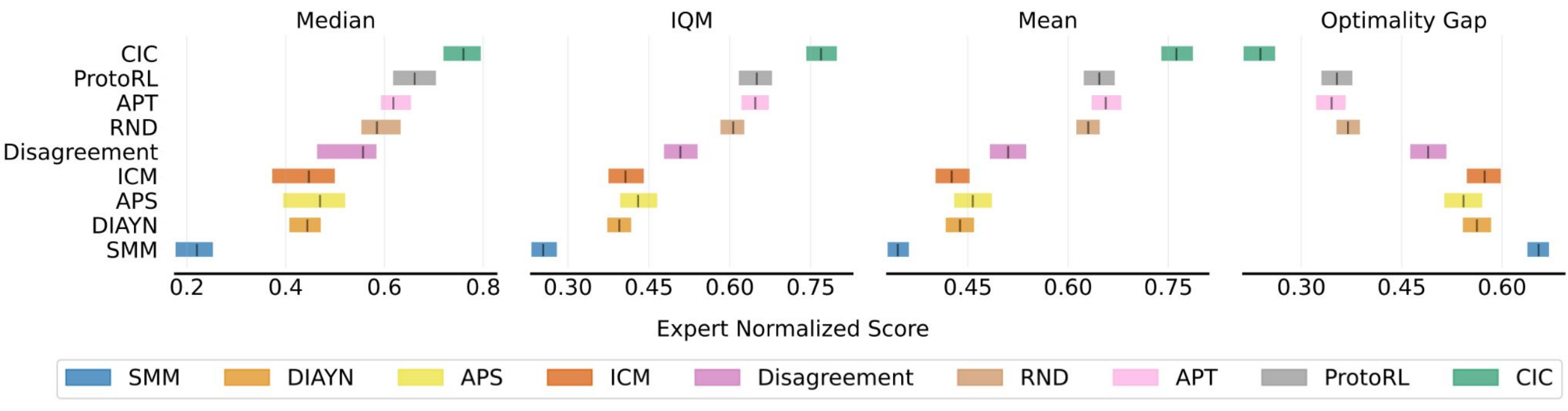}
\caption{ \small We report the aggregate statistics using stratified bootstrap intervals~\citep{agarwal2021rliable} for 12 downstream tasks on URLB with 10 seeds, so each statistic for each algorithm has 120 seeds in total. We find that overall, \name achieves leading performance on URLB in terms of the IQM, mean, and OG statistics. As recommended by~\citet{agarwal2021rliable}, we use the IQM as our primary performance measure. In terms of IQM, \name improves upon the next best skill discovery algorithm (APS) by $79\%$ and the next best algorithm overall (ProtoRL) by $18\%$.}
\label{fig:main}
\vspace{-4mm}
\end{figure*}

\subsection{Contrastive Intrinsic Control}
\label{cic_estimator}

From Section~\ref{sec:problem} we are motivated to find a lower bound for $I(\tau; \skill)$ with a discriminator that is capable of supporting high-dimensional continuous skills\footnote{In high-dimensional state-action spaces the number of distinct behaviors can be quite large.}. Additionally, we wish to increase the diversity of behaviors so that the discriminator can continue learning new skills throughout training. To improve the discriminator, we propose to utilize noise contrastive estimation (NCE)~\citep{gutmann2019nce} between state-transitions and latent skills as a lower bound for $I(\tau;\skill)$.\footnote{Note that $\tau$ is not a trajectory but some function of states.} It has been shown previously that such estimators provide a valid lower bound for mutual information~\cite{oord2018representation}. However, to the best of our knowledge, this is the first work to investigate contrastive representation learning for intrinsic control. 

{\it Representation Learning:} Specifically, we propose to learn embeddings with the following representation learning objective, which is effectively CPC between state-transitions and latent skills:

\begin{align}
    \label{eq:cic}
    I(\tau ; \skill) \geq \expec [f(\tau, z)   - \log \frac 1 N \sum_{j=1}^N \exp (f(\tau_j, z))].
\end{align}
where $f(\tau, \skill)$ is any real valued function. For convenience, we define the discriminator $\log q(\tau|z)$ as
\begin{align}
    \log q(\tau|z)  \coloneqq  f(\tau, z)   & - \log \frac 1 N \sum_{j=1}^N \exp (f(\tau_j, z)).
\end{align}
For our practical algorithm, we parameterize this function as $f(\tau, z) =  g_{\psi_1}(\tau)^\top g_{\psi_2}(z)/{ \| g_{\psi_1} (\tau) \| \| g_{\psi_2} (z) \| T}$ where $\tau = (s,s')$ is a transition tuple, $g_{\psi_k}$ are neural encoders, and $T$ is a temperature parameter. This inner product is similar to the one used in SimCLR~\citep{chen2020simclr}. 

The representation learning loss backpropagates gradients from the NCE loss which maximizes similarity between state-transitions and corresponding skills. 

\begin{align}
F_{NCE}(\tau) & = \frac{g_{\psi_1}(\tau_i)^\top g_{\psi_2}(z_i)}{\| g_{\psi_1} (\tau_i) \| \| g_{\psi_2} (z_i) \| T}    \nonumber \\ - &  \log \frac 1 N \sum_{j=1}^{N} \exp \left ( \frac{g_{\psi_1}(\tau_j)^\top g_{\psi_2}(z_i)}{\| g_{\psi_1} (\tau_j) \| \| g_{\psi_2} (z_i) \| T} \right ) 
\label{eq:cpc_loss}
\end{align}

We provide pseudocode for the CIC representation learning loss below:
\begin{lstlisting}[language=Python, caption=Pseudocode for the CIC loss]
"""
PyTorch-like pseudocode for the CIC loss
"""

def cic_loss(s, s_next, z, temp):
    """
    - states: s, s_next (B, D)
    - skills: z (B, D)
    """
    
    tau = concat(s, s_next, dim=1)
    
    query = query_net(z) 
    key = key_net(tau)
    
    query = normalize(query, dim=1)
    key = normalize(key, dim=1)
    
    """
    positives are on diagonal
    negatives are off diagonal 
    """
    
    logits = matmul(query, key.T) / temp 
    labels = arange(logits.shape[0])
    
    loss = cross_entropy(logits, labels)
    
    return loss
    
 
\end{lstlisting}

{\it Intrinsic reward:} Although we have a representation learning objective, we still need to specify the intrinsic reward for the algorithm for which there can be multiple choices. Prior works consider specifying an intrinsic reward that is proportional to state-transition entropy~\cite{liu2021unsupervised}, the discriminator~\cite{EysenbachGIL19diayn}, a similarity score between states and skills~\cite{farley2021discern}, and the uncertainty of the discriminator~\cite{strouse2021disdain}. We investigate each of these choices and find that an intrinsic reward that maximizes state-transition entropy coupled with representation learning via the CPC loss defined in Sec.~\ref{cic_estimator} is the simplest variant that also performs well (see Table~\ref{tab:int_rew_study}).

\begin{figure*}[t]
\vspace{10mm}
\centering
\begin{subfigure}
    \centering
    \begin{overpic}[width=0.24\textwidth]{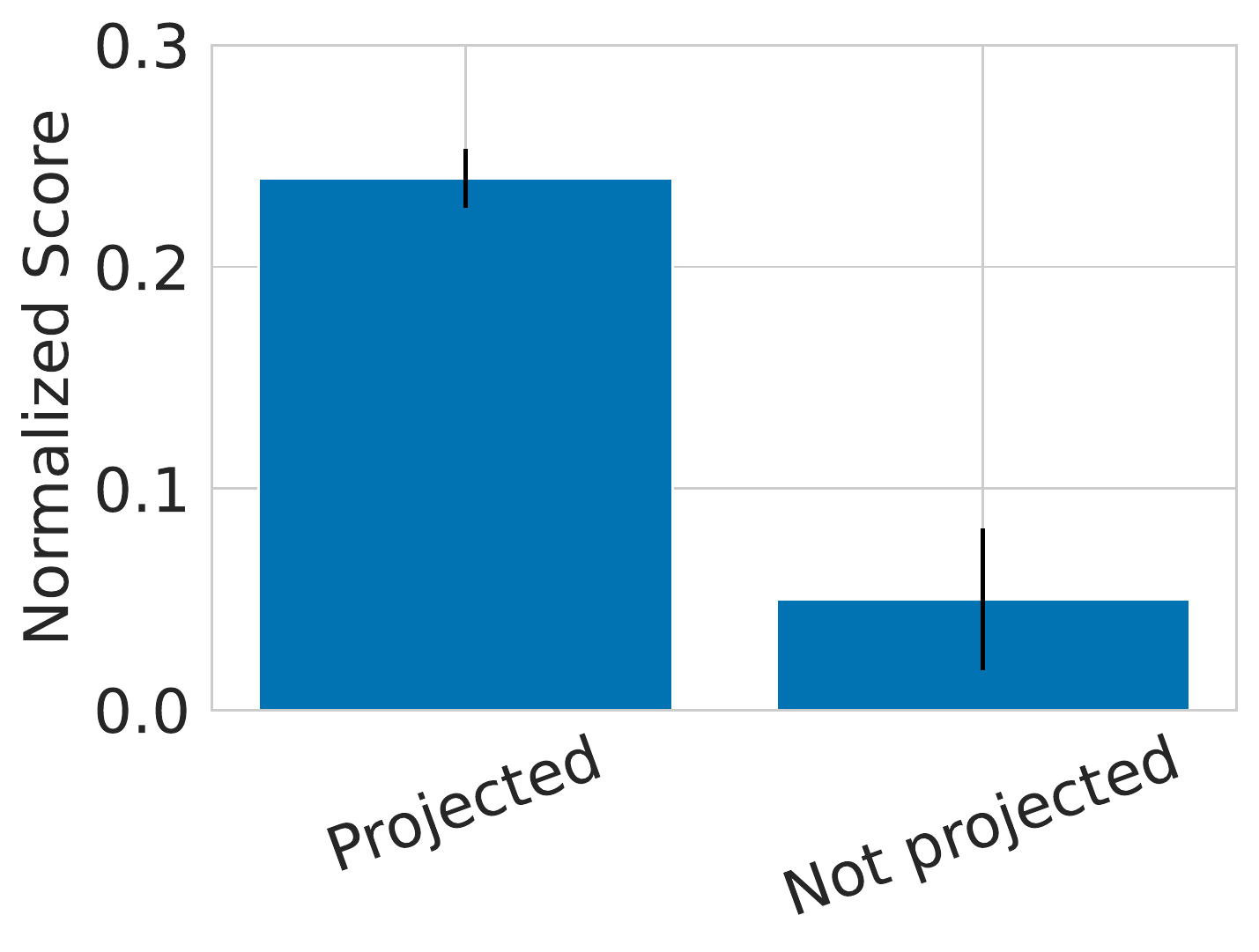}
 \put (10,84) {\textbf{\small(a)}\hspace{0.5em}\small{Skill projection}}
 \end{overpic}
\end{subfigure}
\begin{subfigure}
    \centering
    \begin{overpic}[width=0.24\textwidth]{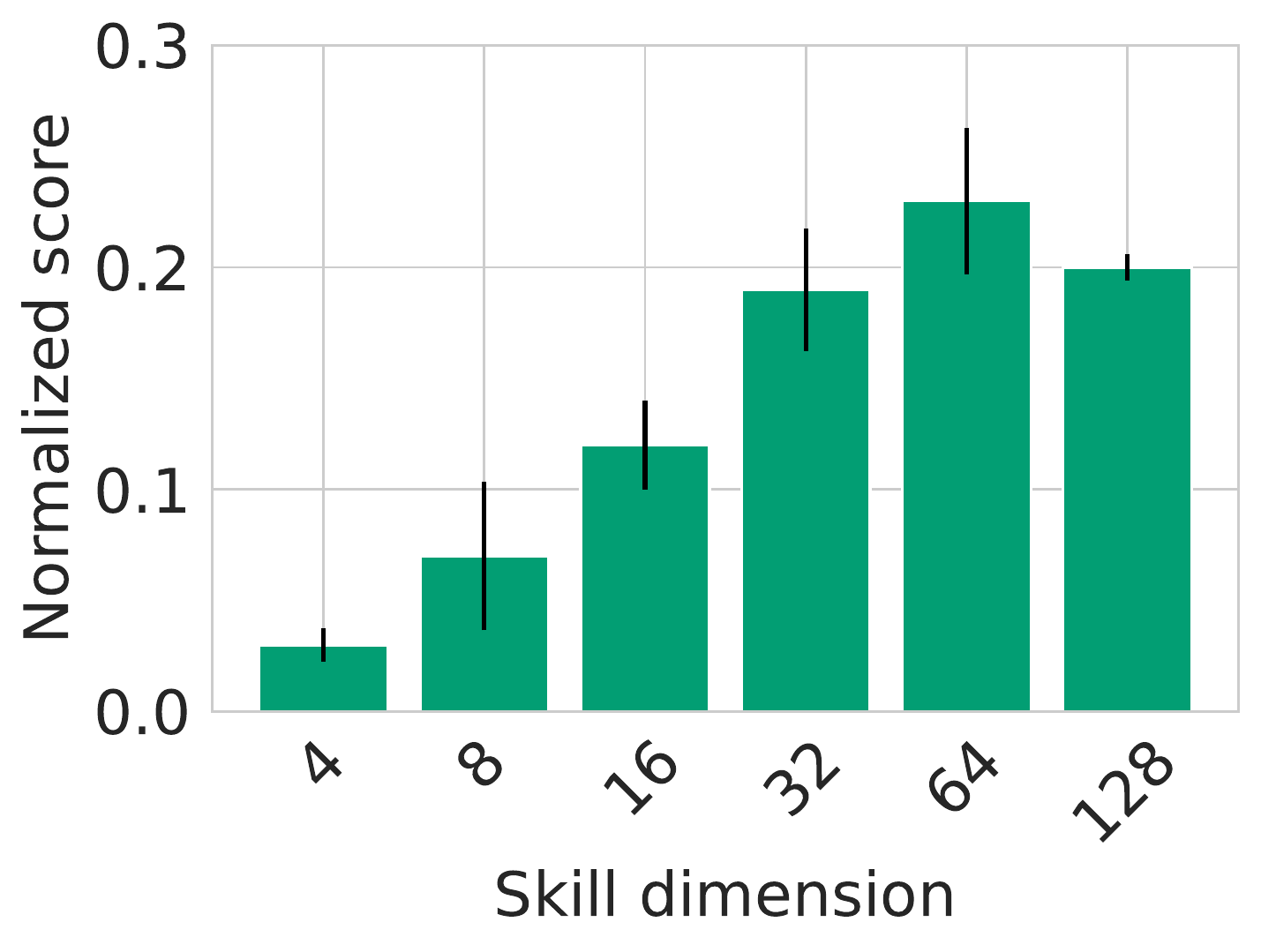}
 \put (10,84) {\textbf{\small(b)}\hspace{0.5em}\small{Skill dimension}}
 \end{overpic}
\end{subfigure}
\begin{subfigure}
    \centering
    \begin{overpic}[width=0.24\textwidth]{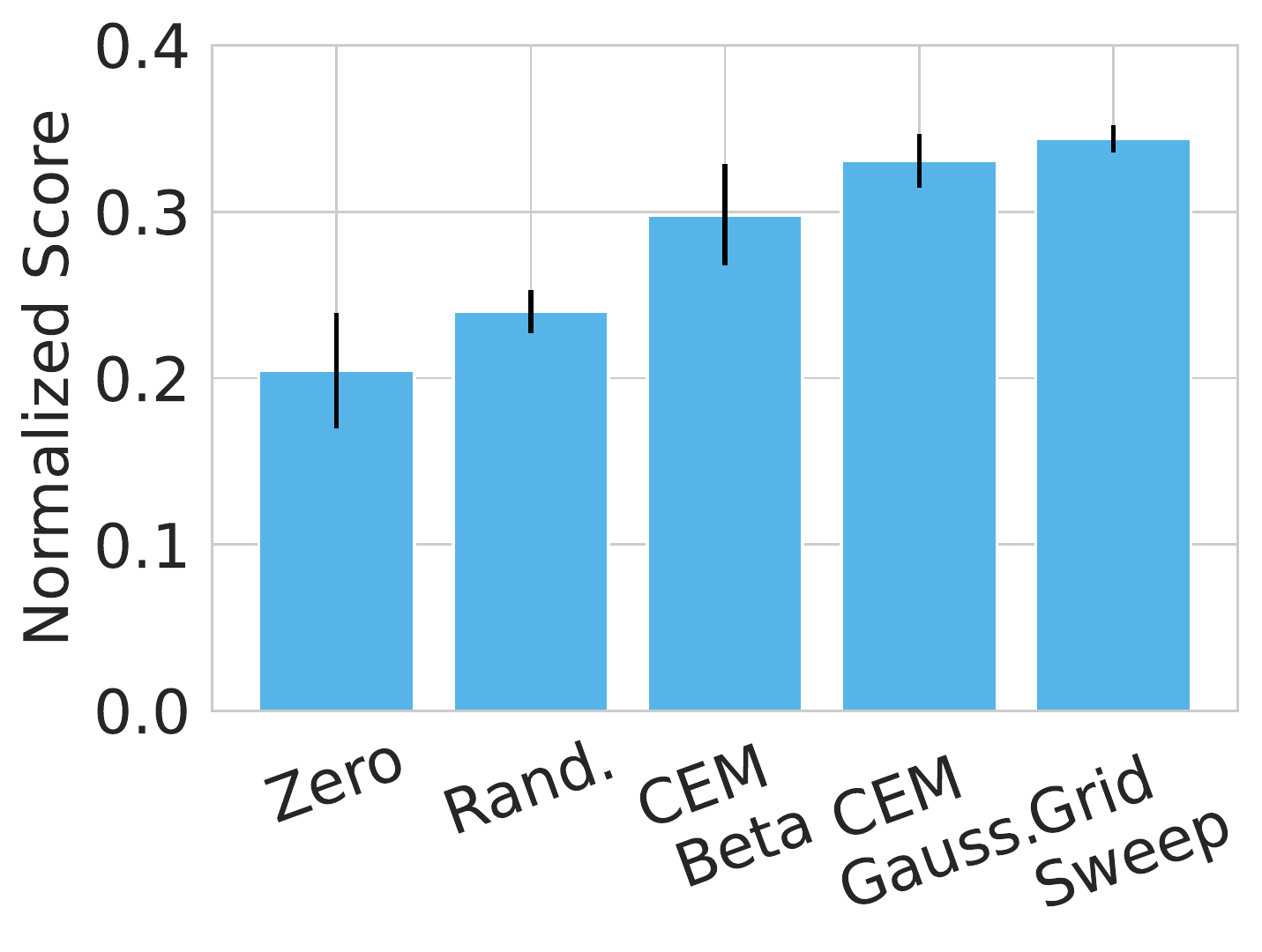}
 \put (10,84) {\textbf{\small(c)}\hspace{0.5em}\small{Skill adaptation}}
 \end{overpic}
\end{subfigure}
\begin{subfigure}
    \centering
    \begin{overpic}[width=0.24\textwidth]{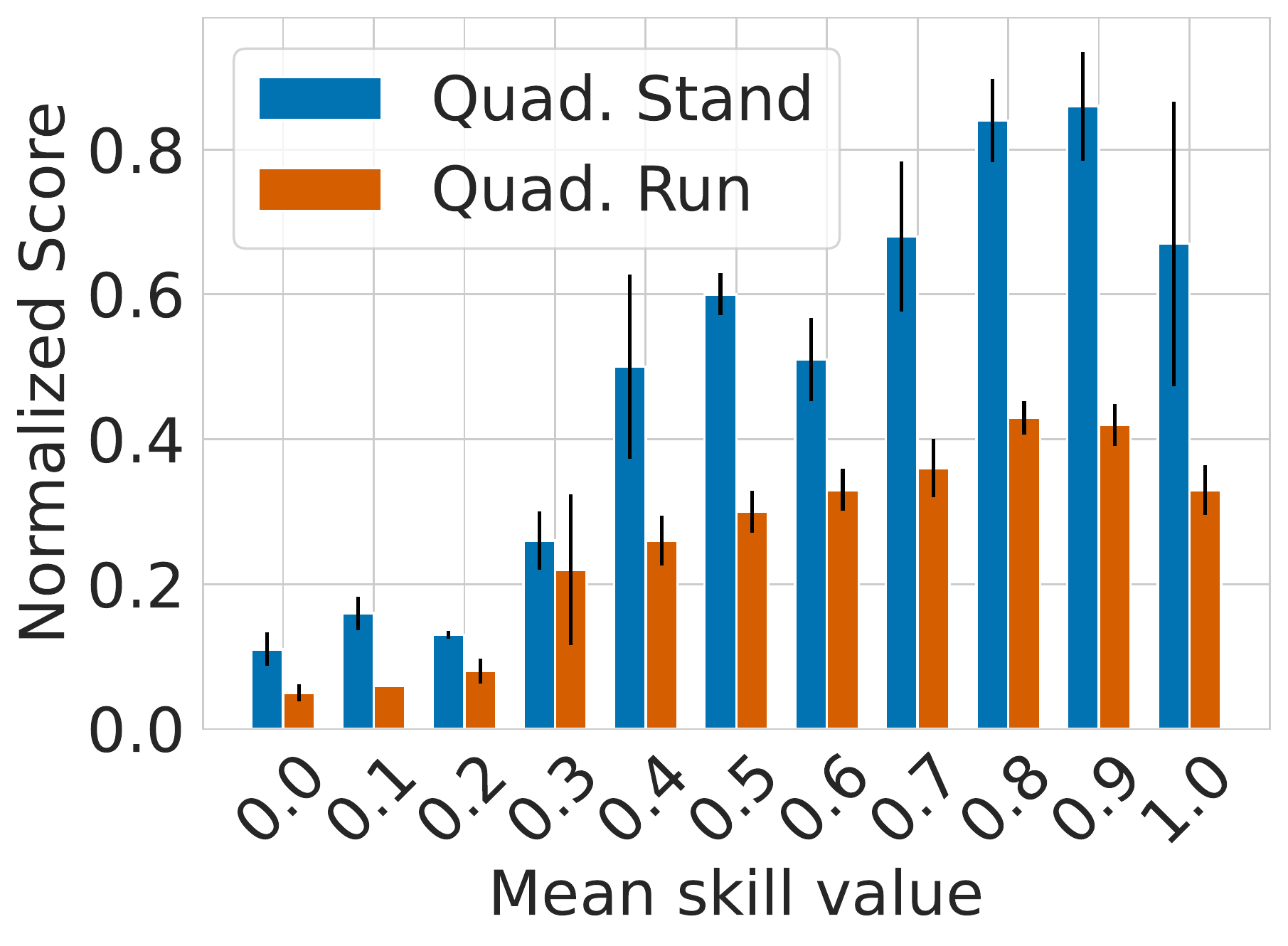}
 \put (10,84) {\textbf{\small(d)}\hspace{0.5em}\small{Skill grid sweep}}
\end{overpic}
\end{subfigure}
\caption{\small Design choices for pre-training and adapting with skills have significant impact on performance. In (a) and (b) the agent's zero-shot performance is evaluated while sampling skills randomly while in (c) and (d) the agent's performance is evaluated after finetuning the skills vector. {\it (a)} we show empirically that the projecting skill vectors after sampling them from noise significantly improves the agent's performance. {\it (b)} The skill dimension is a crucial hyperparameter and, unlike prior methods, \name scales to large skill vectors achieving optimal performance at 64 dimensional skills. {\it (c)} We test several adapation strategies and find that a simple grid search performs best given the small 4k step adaptation budget, {\it (d)} Choosing the right skill vector has substantial impact on performance and grid sweeping allows the agent to select the appropriate skill.    }
\label{fig:skill_abs}
\end{figure*}

For the intrinsic reward, we use a particle estimate~\citep{singh03entropy, beirlant1997nonparametric} as in \citet{liu2021unsupervised} of the state-transition entropy. Similar to ~\citet{liu2021unsupervised, yarats21protorl} we estimate the entropy up to a proportionality constant, because we want the agent to maximize entropy rather than estimate its exact value. 



The APT particle entropy estimate is proportional to the distance between the current visited state transition and previously seen neighboring points. 

\begin{align}
    \mathcal{H}_{particle} (\tau) \propto      \frac 1 {N_k} \sum_{h_{i}^{\star} \in N_k }^{N_k} \log \|h_{i}-h_{i}^{\star}\| 
    \label{eq:apt}
\end{align}

where $h_i$ is an embedding of $\tau_i$ shown in Fig.~\ref{fig:arch}, $h_i^*$ is a kNN embedding, $N_k$ is the number of kNNs, and $N - 1$ is the number of negatives. The total number of elements in the summation is $N$ because it includes one positive.

{\it Explore and Exploit:} With these design choices the two components of the CIC algorithm can be interpreted as {\it exploration} with intrinsic rewards and {\it exploitation} using representation learning to distill behaviors into skills. The marginal entropy maximizes the diversity of state-transition embeddings while the contrastive discriminator $\log q(\tau|z)$ encourages exploitation by ensuring that skills $z$ lead to predictable states $\tau$. Together the two terms incentivize the discovery of diverse yet predictable behaviors from the RL agent. While CIC shares a similar intrinsic reward structure to APT~\cite{liu2021unsupervised}, we show that the new representation learning loss from the CIC estimator results in substantial performance gains in Sec~\ref{sec:results}.

\section{Practical Implementation}
\label{sec:practical}

Our practical implementation consists of two main components: the RL optimization algorithm and the CIC architecture. For fairness and clarity of comparison, we use the same RL optimization algorithm for our method and all baselines in this work. Since the baselines implemented in URLB~\citep{laskin_yarats_2021_urlb} use a DDPG\footnote{It was recently was shown that a DDPG achieves state-of-the-art performance~\citep{yarats2021drqv2} on DeepMind Control~\citep{tassa2018deepmind} and is more stable than SAC~\citep{sac} on this benchmark.}~\citep{lillicrap15ddpg} as their backbone, we opt for the same DDPG architecture to optimize our method as well (see Appendix~\ref{app:ddpg}).

{\it CIC Architecture:} We use a particle estimator as in~\citet{liu2021unsupervised} to estimate $\mathcal H(\tau)$. To compute the variational density $q(\tau|z)$, we first sample skills from uniform noise $z \sim p(z)$ where $p(z)$ is the uniform distribution over the $[0,1]$ interval. We then use two MLP encoders to embed $g_{\psi_1}(\tau)$ and $g_{\psi_2}(z)$, and optimize the parameters $\psi_1, \psi_2$ with the CPC loss similar to SimCLR~\citep{chen2020simclr} since $f(\tau,z) = g_{\psi_1}(\tau)^T g_{\psi_2}(z)$. We fix the hyperparameters across all domains and downstream tasks. We refer the reader to the Appendices~\ref{app:fullcicalgo} and~\ref{appendix:params} for the full algorithm and a full list of hyperparameters.

{\it Adapting to downstream tasks: } To adapt to downstream tasks we follow the same procedure for competence-based method adaptation as in URLB~\citep{laskin_yarats_2021_urlb}. During the first 4k environment interactions we populate the DDPG replay buffer with samples and use the extrinsic rewards collected during this period to finetune the skill vector $z$. While it's common to finetune skills with Cross Entropy Adaptation (CMA), given our limited  budget of 4k samples (only 4 episodes) we find that a simple grid sweep of skills over the interval $[0,1]$ produces the best results (see Fig.~\ref{fig:skill_abs}). After this, we fix the skill $z$ and finetune the DDPG actor-critic parameters against the extrinsic reward for the remaining 96k steps. Note that competence-based methods in URLB also finetune their skills during the first 4k finetuning steps ensuring a fair comparison between the methods. The full adaptation procedure is detailed in Appendix~\ref{app:fullcicalgo}.

 


\section{Experimental Setup}
\label{sec:results}

{\bf Environments} We evaluate our approach on tasks from URLB, which consists of twelve downstream tasks across three challenging continuous control domains for exploration algorithms -- walker, quadruped, and Jaco arm. Walker requires a biped constrained to a 2D vertical plane to perform locomotion tasks while balancing. Quadruped is more challenging due to a higher-dimensional state-action space and requires a quadruped to in a 3D environment to learn locomotion skills. Jaco arm is a 6-DOF robotic arm with a three-finger gripper to move and manipulate objects without locking. All three environments are challenging in the absence of an extrinsic reward.

{\bf Baselines:} We compare \name to baselines across all three exploration categories. Knowledge-based basedlines include ICM~\citep{pathak2017curiosity}, Disagreement~\citep{Pathak19disagreement}, and RND~\citep{burda2018exploration}. Data-based baselines incude APT~\citep{liu2021unsupervised} and ProtoRL~\citep{yarats21protorl}. Competence-based baselines include DIAYN~\cite{EysenbachGIL19diayn}, SMM~\cite{lee2019smm}, and APS~\citep{liu21aps}. The closest baselines to \name are APT, which is similar to \name but without  state-skill CPC representation learning (no discriminator), and APS which uses the same decomposition of mutual information as \name and also uses a particle entropy estimate for $\mathcal H(\tau)$. The main difference between APS and \name is that APS uses successor features while \name uses a contrastive estimator for the discriminator. For further details regarding baselines we refer the reader to Appendix~\ref{appendix:baselines}.

{\bf Evaluation:} We follow an identical evaluation to the 2M pre-training setup in URLB. First, we pre-train each RL agent with the intrinsic rewards for 2M steps. Then, we finetune  each agent to the downstream task with extrinsic rewards for 100k steps. All baselines were run for 10 seeds per downstream task for each algorithm using the code and hyperparameters provided by URLB~\cite{laskin_yarats_2021_urlb}. Built on top of URLB, CIC is also run for 10 seeds per task. A total of $1080 = 9 \text{ algorithms} \times 12 \text{ tasks} \times 10 \text{ seeds}$ experiments were run for the main results. Importantly, all baselines and CIC use a DDPG agent as their backbone. 

To ensure that our evaluation statistics are unbiased we use stratified bootstrap confidence intervals to report aggregate statistics across $M$ runs with $N$ seeds as described in {\it Rliable}~\citep{agarwal2021rliable} to report statistics for our main results in Fig.~\ref{fig:main}. Our primary success metric is the interquartile mean (IQM) and the Optimality Gap (OG). IQM discards the top and bottom 25$\%$ of runs and then computes the mean. It is less susceptible to outliers than the mean and was shown to be the most reliable statistic for reporting results for RL experiments in \citet{agarwal2021rliable}. OG measures how far a policy is from optimal (expert) performance. To define expert performance we use the convention in URLB, which is the score achieved by a randomly initialized DDPG after 2M steps of finetuning (20x more steps than our finetuning budget).

\section{Results}

We investigate empirical answers to the following research questions: (Q1) How does CIC adaptation efficiency compare to prior competence-based algorithms and exploration algorithms more broadly? (Q2) Which intrinsic reward instantiation of CIC performs best? (Q3) How do the two terms in the CIC objective affect algorithm performance? (Q4) How does skill selection affect the quality of the pre-trained policy? (Q5) Which architecture details matter most?

{\bf Adaptation efficiency of CIC and exploration baslines:} Expert normalized scores of CIC and exploration algorithms from URLB are shown in Fig.~\ref{eq:cic}. We find that CIC substantially outperforms prior competence-based algorithms (DIAYN, SMM, APS) achieving a $79\%$ higher IQM than the next best competence-based method (APS) and, more broadly,  achieving a $18\%$  higher IQM than the next best overall baseline (ProtoRL). In further ablations, we find that the contributing factors to CIC's performance are its ability to accommodate  substantially larger continuous skill spaces than prior competence-based methods.

{\bf Intrinsic reward specification:} The intrinsic reward for competence-based algorithms can be instantiated in many different ways. Here, we analyze intrinsic reward for CIC with the form $r_{int} = H(\tau) + D(\tau, z)$, where $D$ is some function of $(\tau,z)$. Prior works, select $D$ to be (i) the discriminator~\cite{liu21aps}, (ii) a cosine similarity between embeddings~\cite{farley2021discern}, (iii) uncertainty of the discriminator~\cite{strouse2021disdain}, and (iv) just the entropy $D(\tau,z) = 0$~\cite{liu2021unsupervised}. We run CIC with each of these variants on the walker and quadruped tasks and measure the final mean performance across the downstream tasks (see Tab.~\ref{tab:int_rew_study}). The results show that the entropy-only intrinsic reward performs best. For this reason the intrinsic reward and representation learning aspects of CIC are decoupled as shown in Eq.~\ref{eq:rewint}. We hypothesize that the reason why a simple entropy-only intrinsic reward works well is that state-skill CPC representation learning clusters similar behaviors together. Since redundant behaviors are clustered, maximizing the entropy of state-transition embeddings produces increasingly diverse behaviors.

\begin{table}[h]
    \centering
    \begin{tabular}{lcccc}
        \toprule
         	 & disc. & similarity &	uncertainty &	entropy \\
         	\midrule
        walker 	& 0.80 & 0.79		 &	0.78	& 0.82 \\
        quad. &	0.44 &	0.63 	 &	0.75 &0.74 \\
        \midrule
        mean & 0.62 & 0.71	 & 0.77 &	0.78 \\
        
        \bottomrule
        \end{tabular}
    \caption{Analyzing four different intrinsic reward specifications for CIC, we find that entropy-based intrinsic reward performs best, suggesting that the CIC discriminator is primarily useful for representation learning. These are normalized scores averaged over 3 seeds across 8 downstream tasks (24 runs per data point).}
    \label{tab:int_rew_study}
\end{table}

\begin{figure}
    \centering
    \includegraphics[width=.9\columnwidth]{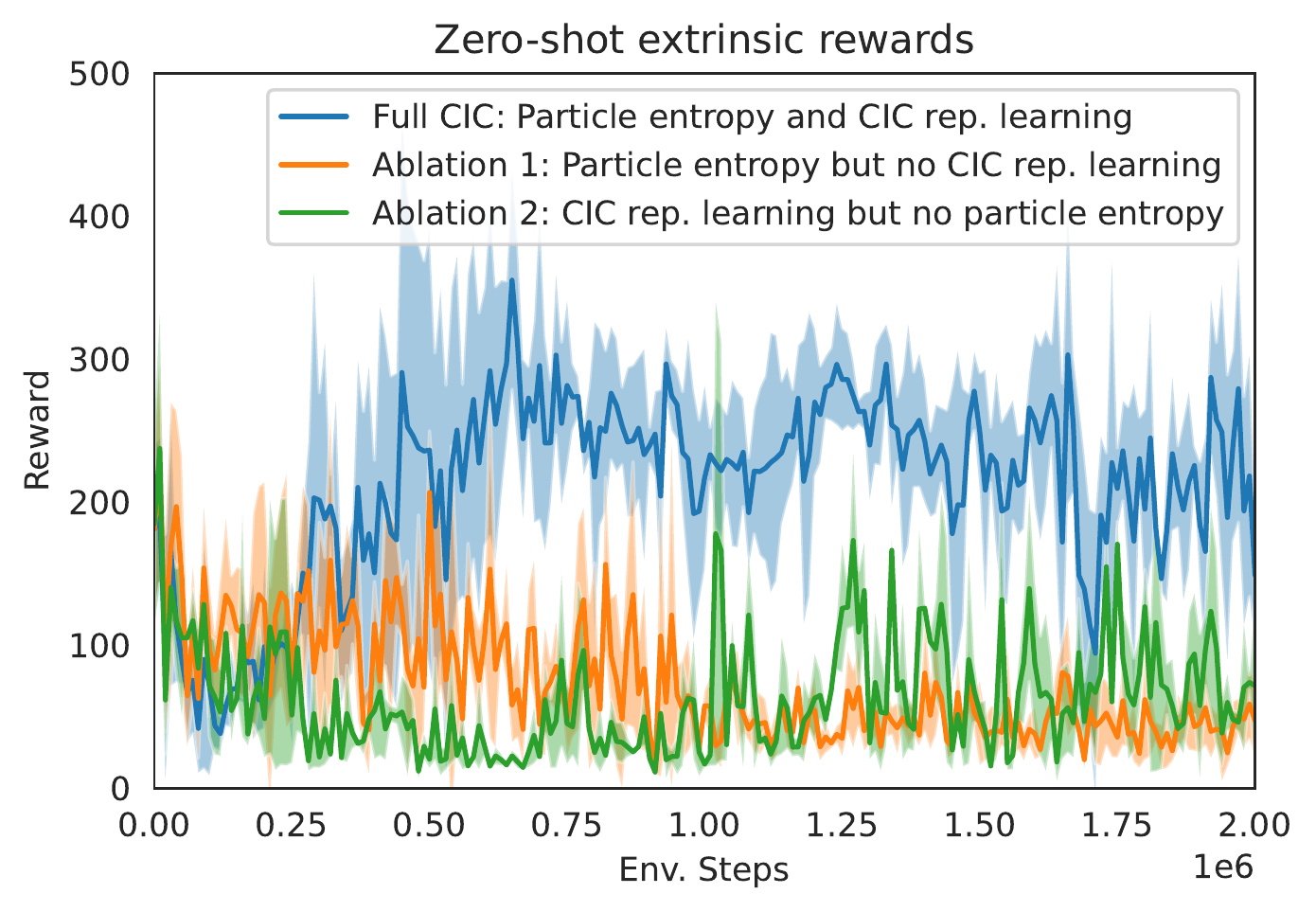}
    \caption{Mean zero-shot extrinsic rewards for Quadruped stand over 3 seeds with and without state-skill representation learning. Without representation learning, the algorithm collapses. Similarly, with CIC representation learning but no entropy term (in which case we use the discriminator as the intrinsic reward) the policy also collapses. Note that there is no finetuning happening here. We're showing the task-specific extrinsic reward during reward-free pre-training as a way to sense-check exploration policy. 
    }
    \label{fig:rep_learning}
    \vspace{-5mm}
\end{figure}

{\bf The importance of representation learning:} To what extent does representation learning with state-skill CPC (see Eq.~\ref{eq:cic}) affect the agent's exploration capability? To answer this question we train the CIC agent with the entropy intrinsic reward with and without the representation learning auxiliary loss for 2M steps. The zero-shot reward plotted in Fig.~\ref{fig:rep_learning} indicates that without representation learning the policy collapses. With representation learning, the agent is able to discover diverse skills evidenced by the non-zero reward. This result suggests that state-skill CPC representation learning is a critical part of CIC.

{\bf  Qualitative analysis of \name behaviors:} Qualitatively, we find that \name is able to learn locomotion behaviors in DMC without extrinsic information such as early termination as in OpenAI Gym. While most skills are higher entropy and thus more chaotic, we show in Fig~\ref{fig:qual_skills} that structured behaviors can be isolated by fixing a particular skill vector. For example, in the walker and quadruped domains - balancing, walking, and flipping skills can be isolated. For more qualitative investigations we refer the reader to Appendix~\ref{app:qualskills}.



{\bf Skill architecture and adaptation ablations:} We find that projecting the skill to a latent space before inputting it as the key for the contrastive loss is an important design decision (see Fig.~\ref{fig:skill_abs}a), most likely because this reduces the diversity of the skill vector making the discriminator task simpler. 

We also find empirically that the skill dimension is an important hyperparameter and that larger skills results in better zero-shot performance (see Fig.~\ref{fig:skill_abs}b), which empirically supports the hypothesis posed in Section~\ref{sec:problem} and Appendix~\ref{appendix:gridworld} that larger skill spaces are important for internalizing  diverse behaviors. Interestingly, CIC zero-shot performance is poor in lower skill dimensions (e.g. $\text{dim}(z) < 10$), suggesting that when $\text{dim}(z)$ is small CIC performs no better than prior competence-based methods such as DIAYN, and that scaling to larger skills enables CIC to pre-train effectively.

To measure the effect of skill finetuning described in Section~\ref{sec:practical}, we sweep mean skill values along the interval of the uniform prior $[0,1]$ with a budget of 4k total environment interactions and read out the performance on the downstream task. By sweeping, we mean simply iterating over the interval $[0,1]$ with fixed step size (e.g. $v=0,0.1,\dots,0.9,1)$ and setting $z_i = v$ for all $i$. This is not an optimal skill sampling strategy but works well due to the extremely limited number of samples for skill selection. We evaluate this ablation on the Quadruped Stand and Run downstream tasks. The results shown in Fig.~\ref{fig:skill_abs} indicate that skill selection can substantially affect zero-shot downstream task performance.

\vspace{-3mm}
\section{Conclusion}
\vspace{-2mm}

 We have introduced a new competence-based algorithm -- \namedef-- which enables more effective exploration than prior  unsupervised skill discovery algorithms by explicitly encouraging diverse behavior while distilling predictable behaviors into skills with a contrastive discriminator. We showed that \name is the first competence-based approach to achieve leading performance on URLB. We hope that this encourages further research in developing RL agents capable of generalization.
 
 \section{Acknowledgements}
We would like to thank Ademi Adeniji, Xinyang Geng, Fangchen Liu for helpful discussions. We would also like to thank Phil Bachman for useful feedback. This work was partially supported by Berkeley DeepDrive, NSF AI4OPT AI Institute for Advances in Optimization under NSF 2112533, and the Office of Naval Research grant N00014-21-1-2769.

\bibliography{main}
\bibliographystyle{icml2022}

\newpage
\appendix

\onecolumn

\appendix

\section{Competence-based Exploration Algorithms}

The competence-based algorithms considered in this work aim to maximize $I(\tau;s)$. The algorithms differ by ho they decompose mutual information, whether they explicitly maximize behavioral entropy, their skill space (discrete or continuous) and their intrinsic reward structure. We provide a list of common competence-based algorithms in Table~\ref{table:skill_algos}.

\begin{table}[h]
\caption{Prior Competence-based Unsupervised Skill Discovery Algorithms }
\label{table:skill_algos}
\begin{center}
\resizebox{\textwidth}{!}{\begin{tabular}{llccll}
\\
\hline
\multicolumn{1}{c}{ Algorithm}  &\multicolumn{1}{c}{ Intrinsic Reward}  & \multicolumn{1}{c}{ Decomposition}  &\multicolumn{1}{c}{Explicit $\max \mathcal H(\tau)$  } &  Skill Dim. & Skill Space

\\ \hline \\
SSN4HRL~\citep{florensa2017stochastic}         & $\log q_\psi (z|s_t)$  & $H(\skill) - H(\skill |\tau)$ & No &  6  & discrete \\
VIC~\citep{GregorRW17vic}         & $ \log q_\psi (z|s_H) )$ & $H(\skill) - H(\skill |\tau)$ & No & 60 & discrete  \\
VALOR~\citep{achiam2018valor}         & $\log q_\psi (z|s_{1:H})$ & $H(\skill) - H(\skill |\tau)$ & No & 64 & discrete \\
DIAYN~\citep{EysenbachGIL19diayn}          & $\log q_\psi (z|s_t) $ & $H(\skill) - H(\skill |\tau)$ & No & 128  & discrete  \\
DADS~\citep{SharmaGLKH20dads}        & $ q_\psi (s'|z, s) - \sum_i \log q(s'|z_i, s)$ & $H(\tau) - H(\tau |z)$ & Yes & 5 & continuous \\
VISR~\citep{hansen20visr}         & $\log q_\psi (z|s_t) $ & $H(\skill) - H(\skill |\tau)$ & No & 10 & continuous  \\
APS~\citep{liu21aps}      &  $ F_\text{Successor} (s|z) + \mathcal H_\text{particle} (s)$ &$\mathcal H(\tau) - \mathcal H(\tau|z)$ &Yes & 10 & continuous \\
\end{tabular}}

\end{center}
\caption{\small A list of competence-based algorithms. We describe the intrinsic reward optimized by each method and the decomposition of the mutual information utilized by the method. We also note whether the method explicitly maximizes state transition entropy. Finally, we note the maximal dimension used in each work and whether the skills are discrete or continuous. All methods prior to \name only support small skill spaces, either because they are discrete or continuous but low-dimensional.   }

\end{table}

\section{Deep Deterministic Policy Gradient (DDPG)}
\label{app:ddpg}

A DDPG is an actor-critic RL algorithm that performs off-policy gradient updates and learns a Q function $Q_\phi (\state, \action)$ and an actor $\pi_\theta (\action | \state)$. The critic is trained by satisfying the Bellman equation.
\begin{equation}
    \label{eq:critic}
    \mathcal{L}_Q(\phi,\mathcal{D})=\mathbb{E}_{(\state_t, \action_t, \reward_t, \state_{t+1})\sim\mathcal{D}}\left[\Big(Q_\phi(\state_t,\action_t) - \reward_t - \gamma Q_{\bar{\phi}}(\state_{t+1},\pi_\theta(\state_{t+1})\Big)^2\right].
\end{equation}
Here, $\bar{\phi}$ is the Polyak average of the parameters $\phi$. As the critic minimizes the Bellman error, the actor maximizes the action-value function.
\begin{equation}
    \label{eq:actor}
    \mathcal{L}_\pi(\theta, \mathcal{D})=\mathbb{E}_{\state_t \sim\mathcal{D}}\left[Q_\phi(\state_t,\pi_\theta(\state_t))\right].
\end{equation}

\section{Baselines}
\label{appendix:baselines}

For baselines, we choose the existing set of benchmarked unsupervised RL algorithms on URLB. We provide a quick summary of each method. For more detailed descriptions of each baseline we refer the reader to URLB~\citep{laskin_yarats_2021_urlb}

{\it Competence-based Baselines:} \name is a competence-based exploration algorithm. For baselines, we compare it to DIAYN~\citep{EysenbachGIL19diayn}, SMM~\citep{lee2019smm}, and APS~\citep{liu21aps}. Each of these algorithms is described in Table~\ref{table:skill_algos}. Notably, APS is a recent state-of-the-art competence-based method that is the most closely related algorithm to the \name algorithm.

{\it Knowledge-based Baselines:} For knowledge-based baselines, we compare to ICM~\cite{pathak2017curiosity}, Disagreement~\cite{Pathak19disagreement}, and RND~\cite{burda2018exploration}. ICM and RND train a dynamics model and random network prediction model and define the intrinsic reward to be proportional to the prediction error. Disagreement trains an ensemble of dynamics models and defines the intrinsic reward to be proportional to the uncertainty of an ensemble.

{\it Data-based Baselines:} For data-based baselines we compare to APT~\citep{liu2021unsupervised} and ProtoRL~\citep{yarats21protorl}. Both methods use a particle estimator to estimate the state visitation entropy. ProtoRL also performs discrete contrastive clustering as in~\citet{caron20swav} as an auxiliary task and uses the resulting clusters to compute the particle entropy. While ProtoRL is more effective than APT when learning from pixels, on state-based URLB APT is competitive with ProtoRL. Our method \name is effectively a skill-conditioned APT agent with a contrastive discriminator.

\section{Relation to Prior Skill Discovery Methods}
\label{appendix:prior_skills}
The most closely relatd prior algorithms to \name are APT~\cite{liu2021unsupervised} and APS~\cite{liu21aps}. Both \name and APS use the $\mathcal H(\tau) - \mathcal H(\tau|z)$ decomposition of the mutual information and both used a particle estimator~\citep{singh03entropy} to compute the state entropy as in~\citet{liu2021unsupervised}. The main difference between \name and APS is the discriminator. APS uses successor features as in~\citet{hansen20visr} for its discriminator while \name uses a noise contrastive estimator. Unlike successor features, which empirically only accommodate low-dimensional continuous skill spaces (see Table~\ref{table:skill_algos}), the noise contrastive discriminator is able to leverage higher continuous dimensional skill vectors. Like APT, \name has an intrinsic reward that maximizes $\mathcal H (\tau)$. However, CIC also does contrastive skill learning to shape the embedding space and outputs a skill-conditioned policy.

The \name discriminator is similar to the one used in DISCERN~\citep{farley2021discern}, a goal-conditioned unsupervised RL algorithm. Both methods use a contrastive discriminator by sampling negatives and computing an inner product between queries and keys. The main differences are (i) that DISCERN maximizes $I(\tau;g)$ where $g$ are image goal embeddings while \name maximizes $I(\tau;z)$ where $z$ are abstract skill vectors; (ii) DISCERN uses the DIAYN-style decomposition $I(\tau;g) = H(g) - H(g| \tau)$ while \name decomposes through $H(\tau) - H(\tau|z)$, and (iii) DISCERN discards the $H(g)$ term by sampling goals uniformly while \name explicitly maximizes $\mathcal H(\tau)$. While DISCERN and \name share similarities, DISCERN operates over image goals while \name operates over abstract skill vectors so the two methods are not directly comparable.

Finally, another similar algorithm to \name is DADS~\citep{SharmaGLKH20dads} which also decomposes through $H(\tau) - H(\tau|z)$. While \name uses a contrastive density estimate for the discriminator, DADS uses a maximum likelihood estimator similar to DIAYN. DADS maximizes $I(s'|s,z)$ and estimates entropy $\mathcal H (s'|s)$ by marginalizing over $z$ such that $\mathcal H(s'|s) = - \log \sum_i q(s' |s, z_i)$ while \name uses a particle estimator.

\section{Full \name Algorithm}
\label{app:fullcicalgo}

The full \name algorithm with both pre-training and fine-tuning phases is shown in Algorithm~\ref{alg:cicfull}. We pre-train \name for 2M steps, and finetune it on each task for 100k steps.

\begin{algorithm}[h]
\caption{\texttt{Contrastive Intrinsic Control}} \label{alg:cicfull}
\begin{algorithmic}[1]
\footnotesize
\Require Initialize all networks: encoders $g_{\psi_1}$ and $g_{\psi_2}$, actor $\pi_\theta$, critic $Q_\phi$,  replay buffer $\mathcal{D}$.
\Require Environment (env), $M$ downstream tasks $T_k, k\in [1,\dots,M]$.
\Require pre-train $N_{\mathrm{PT}}=2M$ and fine-tune $N_{\mathrm{FT}}=100K$ steps.

\For{$t = 1..N_{\mathrm{PT}}$} \Comment{Part 1: Unsupervised Pre-training}
\State Sample and encode skill $z \sim p(z)$ and $z \leftarrow g_{\psi_2} (z)$
\State  Encode state $s_t \leftarrow g_{\psi_1}(s_t)$ and sample action $\action_t \leftarrow \pi_\theta( s_{t},z) + \epsilon$ where $\epsilon \sim \mathcal{N}(0, \sigma^2)$  
\State  Observe next state $s_{t+1} \sim P(\cdot | s_{t}, \action_t)$ 
\State Add transition to replay buffer $\mathcal{D} \leftarrow \mathcal{D} \cup (s_t,  a_t, s_{t+1})$
\State Sample a minibatch from $\mathcal{D}$, compute contrastive loss in Eq.\ref{eq:cic} and update encoders $g_{\psi_1}, g_{\psi_2}$, compute \name intrinsic reward with Eq.~\ref{eq:rewint} and update actor $\pi_\theta$ and critic $Q_\phi$
\EndFor

\For{$T_k \in [T_1,\dots,T_M$]} \Comment{Part 2: Supervised Fine-tuning}

\State Initialize all networks with weights from pre-training phase and an empty replay buffer $\mathcal{D}$.
\For{$t = 1\dots4,000$}
\State Take random action $a_t \sim \mathcal{N} (0, 1)$
\State Select skill with grid sweep over unit interval $[0,1]$ every 100 steps
\State Sample minibatch from $\mathcal {D}$ and update actor $\pi_\theta$ and critic $Q_\phi$

\EndFor
\State Fix skill $z$ that achieved highest extrinsic reward during grid sweep.
\For{$t = 4,000\dots N_{\mathrm{FT}}$}
\State Encode state $s_t \leftarrow g_{\psi_1}(s_t)$ and sample action $a_t \leftarrow \pi_\theta( s_{t},z) + \epsilon$ where $\epsilon \sim \mathcal{N}(0, \sigma^2)$  
\State  Observe next state and reward $s_{t+1} , \reward^{\mathrm{ext}}_t \sim P(\cdot | s_{t}, \action_t)$ 
\State Add transition to replay buffer $\mathcal{D} \leftarrow \mathcal{D} \cup (s_t,  a_t, \reward^{\mathrm{ext}}_t , s_{t+1})$
\State Sample minibatch from $\mathcal{D}$ and update actor $\pi_\theta$ and critic $Q_\phi$.
\EndFor
\State Evaluate performance of RL agent on task $T_k$
\EndFor

\end{algorithmic}
\end{algorithm}

\section{Hyper-parameters}
\label{appendix:params}

Baseline hyperparameters are taken from URLB~\cite{laskin_yarats_2021_urlb}, which were selected by performing a grid sweep over tasks and picking the best performing set of hyperparameters. Except for the skill dimension, hyperparameters for \name are borrowed from URLB.

\begin{table}[h]
\caption{\label{table:common_hyper_params} Hyper-parameters used for \name. }
\centering
\begin{tabular}{lc}
\hline
DDPG hyper-parameter       & Value \\
\hline
\: Replay buffer capacity & $10^6$ \\
\: Action repeat & $1$ \\
\: Seed frames & $4000$ \\
\: $n$-step returns & $3$ \\
\: Mini-batch size & $1024$  \\
\: Seed frames & $4000$ \\
\: Discount ($\gamma$) & $0.99$ \\
\: Optimizer & Adam \\
\: Learning rate & $10^{-4}$ \\
\: Agent update frequency & $2$ \\
\: Critic target EMA rate ($\tau_Q$) & $0.01$ \\
\: Features dim. & $1024$  \\
\: Hidden dim. & $1024$ \\
\: Exploration stddev clip & $0.3$ \\
\: Exploration stddev value & $0.2$ \\
\: Number pre-training frames &  $2\times 10^6$ \\
\: Number fine-turning frames & $1 \times 10^5$ \\
\hline
\name hyper-parameter       & Value \\
\hline

\: Skill dim & 64 continuous \\
\: Prior & Uniform [0,1] \\
\: Skill sampling frequency (steps) & 50 \\
\: State net arch. $g_{\psi_1}(s)$ & $ \dim (\mathcal O) \to 1024 \to 1024 \to 64$ $\textrm{ReLU}$ MLP  \\
\: Skill net arch. $g_{\psi_2}(z)$ & $64 \to 1024 \to 1024 \to 64$ $\textrm{ReLU}$ MLP  \\
\: Prediction net arch. & $64 \to 1024 \to 1024 \to 64$ $\textrm{ReLU}$ MLP  \\

\hline
\end{tabular}

\end{table}

\section{Raw Numerical Results}

\begin{table*}
\centering
\begin{tabular}{lccccccccccc}
\toprule
 Statistic & ICM &  Dis. &  RND &  APT &  Proto &  DIAYN &  APS &  SMM &  CIC & $\%$ CIC $>$ APS & $\%$ CIC $>$ Proto \\
 
\midrule
Median $\uparrow$ & 0.45 &          0.56 & 0.58 & 0.62  & 0.66 &   0.44 & 0.47 & 0.22 & 0.76 & {\color{mygreen}+61$\%$} & {\color{mygreen}+15$\%$ }\\

IQM $\uparrow$  & 0.41 &          0.51 & 0.61 & 0.65 &     0.65 &   0.40 & 0.43 & 0.25 & 0.77 & {\color{mygreen}+79$\%$} & {\color{mygreen}+18$\%$}\\

Mean $\uparrow$  & 0.43 &          0.51 & 0.63 & 0.66 &     0.65 &   0.44 & 0.46 & 0.35 & 0.76 & {\color{mygreen}+65$\%$} & {\color{mygreen}+17$\%$} \\
OG $\downarrow$  & 0.57 &          0.49 & 0.37 & 0.35 &     0.35 &   0.56 & 0.54 & 0.65 & 0.24 & {\color{mygreen}-44$\%$} & {\color{mygreen}-68$\%$} \\
\bottomrule
\end{tabular}
\caption{Statics for downstream task normalized scores for CIC and baselines from URLB~\cite{laskin_yarats_2021_urlb}. CIC improves over both the prior leading competence-based method APS~\cite{liu21aps} and overall next-best exploration algorithm ProtoRL~\cite{yarats21protorl} across all readout statistics. Each data point is a statistic computed using 10 seeds and 12 downstream tasks (120 experiments per data point). The statistics are computed using RLiable~\cite{agarwal2021rliable}.} 
    \label{tab:res_table}
\end{table*}

We provide a list of raw numerical results for finetuning \name and baselines in Tables~\ref{tab:res_table} and~\ref{table:states_numbers}. All baselines were run using the code provided by URLB~\cite{laskin_yarats_2021_urlb} for 10 seeds per downstream task.

\begin{table}[h]
\centering
\resizebox{\columnwidth}{!}{
\begin{tabular}{|lc|c|c|c|ccc|cc|ccc|}
\hline
\multicolumn{11}{c}{Pre-trainining for $2 \times 10^6$ environment steps}\\
\hline
Domain & Task & Expert & DDPG & CIC & ICM & Disagreement & RND & APT & ProtoRL & SMM & DIAYN & APS \\
\hline
\multirow{4}{*}{Walker}&  Flip & 799 &  538$\pm$27 &  631 $\pm$ 34& 417$\pm$16 & 346$\pm$13 & 474$\pm$39 & 544$\pm$14 & 456$\pm$12 & 450$\pm$24 & 319$\pm$17 & 465$\pm$20\\
 &  Run & 796& 325$\pm$25 &  486 $ \pm$  25 &  247$\pm$21 & 208$\pm$15 & 406$\pm$30 & 392$\pm$26 & 306$\pm$13 &  426$\pm$26 & 158$\pm$8 & 134$\pm$16\\
 &  Stand & 984& 899$\pm$23 &  959 $ \pm $ 2 & 859$\pm$23 & 746$\pm$34 & 911$\pm$5 &  942$\pm$6  & 917$\pm$27 & 924$\pm$12 & 695$\pm$46 & 721$\pm$44\\
 &  Walk & 971& 748$\pm$47 &  885 $\pm$  28  & 627$\pm$42 & 549$\pm$37 & 704$\pm$30 & 773$\pm$70 & 792$\pm$41 & 770$\pm$44 & 498$\pm$27 & 527$\pm$79\\
\hline
\multirow{4}{*}{Quadruped}&  Jump & 888 & 236$\pm$48 &   595 $ \pm$  42 & 178$\pm$35 & 389$\pm$62 & 637$\pm$12 & 648$\pm$18 & 617$\pm$44 & 96$\pm$7 &  660$\pm$43 & 463$\pm$51\\
 &  Run & 888 &157$\pm$31 &  505 $ \pm$  47 & 110$\pm$18 & 337$\pm$30 & 459$\pm$6 &  492$\pm$14 & 373$\pm$33 & 96$\pm$6 & 433$\pm$29 & 281$\pm$17\\
 &  Stand & 920& 392$\pm$73 & 761 $ \pm$  54 &  312$\pm$68 & 512$\pm$89 & 766$\pm$43 &  872$\pm$23 & 716$\pm$56 & 123$\pm$11 &  851$\pm$43  & 542$\pm$53\\
 &  Walk & 866 & 229$\pm$57 &  723 $ \pm$  43 & 126$\pm$27 & 293$\pm$37 & 536$\pm$39 &  770$\pm$47 & 412$\pm$54 & 80$\pm$6 & 576$\pm$81 & 436$\pm$79\\
\hline
\multirow{4}{*}{Jaco}&  Reach bottom left & 193 & 72$\pm$22 &  138 $ \pm$  9 & 111$\pm$11 &   124$\pm$7 & 110$\pm$5 & 103$\pm$8 &  129$\pm$8 & 45$\pm$7 & 39$\pm$6 & 76$\pm$8\\
 &  Reach bottom right & 203 & 117$\pm$18 &  145 $ \pm$  7 & 97$\pm$9 & 115$\pm$10 & 117$\pm$7 & 100$\pm$6 & 132$\pm$8 & 46$\pm$11 & 38$\pm$5 & 88$\pm$11\\
 &  Reach top left & 191 & 116$\pm$22 &  153 $ \pm$  7  & 82$\pm$14 & 106$\pm$12 & 99$\pm$6 & 73$\pm$12 & 123$\pm$9 & 36$\pm$3 & 19$\pm$4 & 68$\pm$6\\
 &  Reach top right & 223 & 94$\pm$18 &  163 $ \pm$  4 &103$\pm$11 & 139$\pm$7 & 100$\pm$6 & 90$\pm$10 & 159$\pm$7 & 47$\pm$6 & 28$\pm$6 & 76$\pm$10\\
\hline
\end{tabular}}
\caption{Performance of \name and baselines on state-based URLB after first pre-training for $2 \times 10^6$ steps and then finetuning with extrinsic rewards for $1 \times 10^5$. All baselines were run for 10 seeds per downstream task for each algorithm using the code provided by URLB~\cite{laskin_yarats_2021_urlb}. A total of $1080 = 9 \text{ algorithms} \times 12 \text{ tasks} \times 10 \text{ seeds}$ experiments were run. }
\label{table:states_numbers}
\end{table}

\newpage 


\section{Toy Example to Illustrate the Need for Larger Skill Spaces}
\label{appendix:gridworld}

\begin{figure*} [h] \centering
\includegraphics[width=\textwidth]{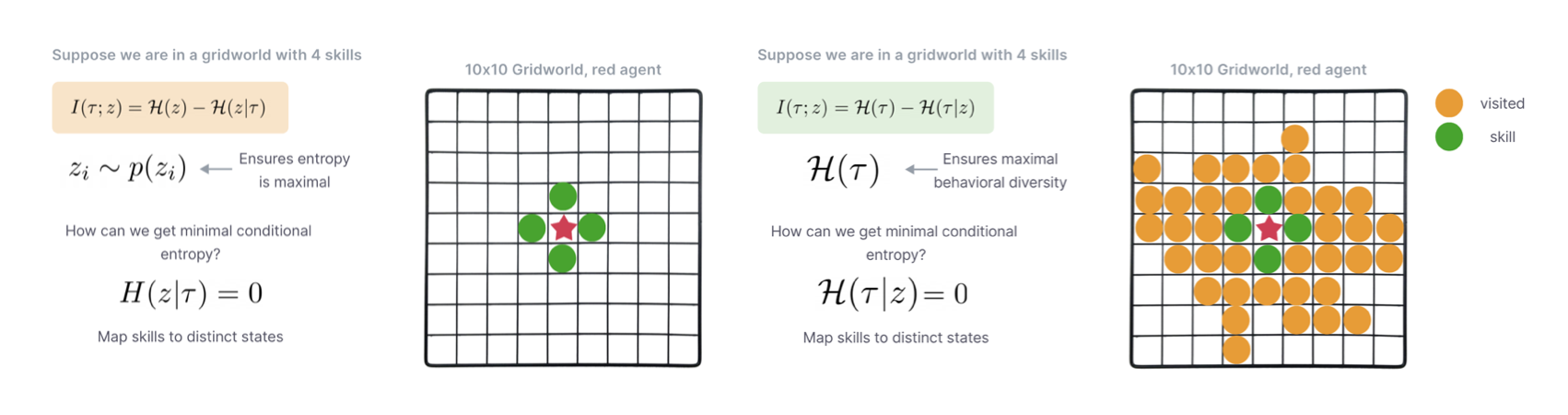}
\caption{\small A gridworld example motivating the need for large skill spaces. In this environment, we place an agent in a $10 \times 10 $ gridworld and provide the agent access to four discrete skills. We show that the mutual information objective can be maximized by mapping these four skills to the nearest neighboring states resulting in low behavioral diversity and exploring only four of the hundred available states.
}
\label{fig:gridworld}
\end{figure*}

We illustrate the need for larger skill spaces with a gridworld example. Suppose we have an agent in a $10\times10$ sized gridworld and that we have four discrete skills at our disposal. Now let $\tau = s$ and consider how we may achieve maximal $I(\tau;z)$ in this setting. If we decompose $I(\tau;z) = \mathcal H(z) - \mathcal H(z|\tau)$ then we can achieve maximal $\mathcal H(z)$ by sampling the four skills uniformly $z \sim p(z)$. We can achieve $\mathcal H(z|\tau) = 0$ by mapping each skill to a distinct neighboring state of the agent. Thus, our mutual information is maximized but as a result the agent only explores four out of the hundrend available states in the gridworld.

Now suppose we consider the second decomposition $I(\tau;z) = \mathcal H(\tau) - \mathcal H(\tau|z) $. Since the agent is maximizing $\mathcal H(\tau) $ it is likely to visit a diverse set of states at first. However, as soon as it learns an accurate discriminator we will have $\mathcal H(\tau|z)$ and again the skills can be mapped to neighboring states to achieve minimal conditional entropy. As a result, the skill conditioned policy will only be able to reach four out of the hundrend possible states in this gridworld. This argument is shown visually in Fig.~\ref{fig:gridworld}.

Skill spaces that are too large can also be an issue. Consider if we had $100$ skills at our disposal in the same gridworld. Then the agent could minimize the conditional entropy by mapping each skill to a unique state which would result in the agent memorizing the environment by finding a one-to-one mapping between states and skills. While this is a potential issue it has not been encountered in practice yet since current competence-based methods support small skill spaces relative to the observation space of the environment.

\section{Qualitative Analysis of Skills}
\label{app:qualskills}

\begin{figure*} [h] \centering
\includegraphics[width=\textwidth]{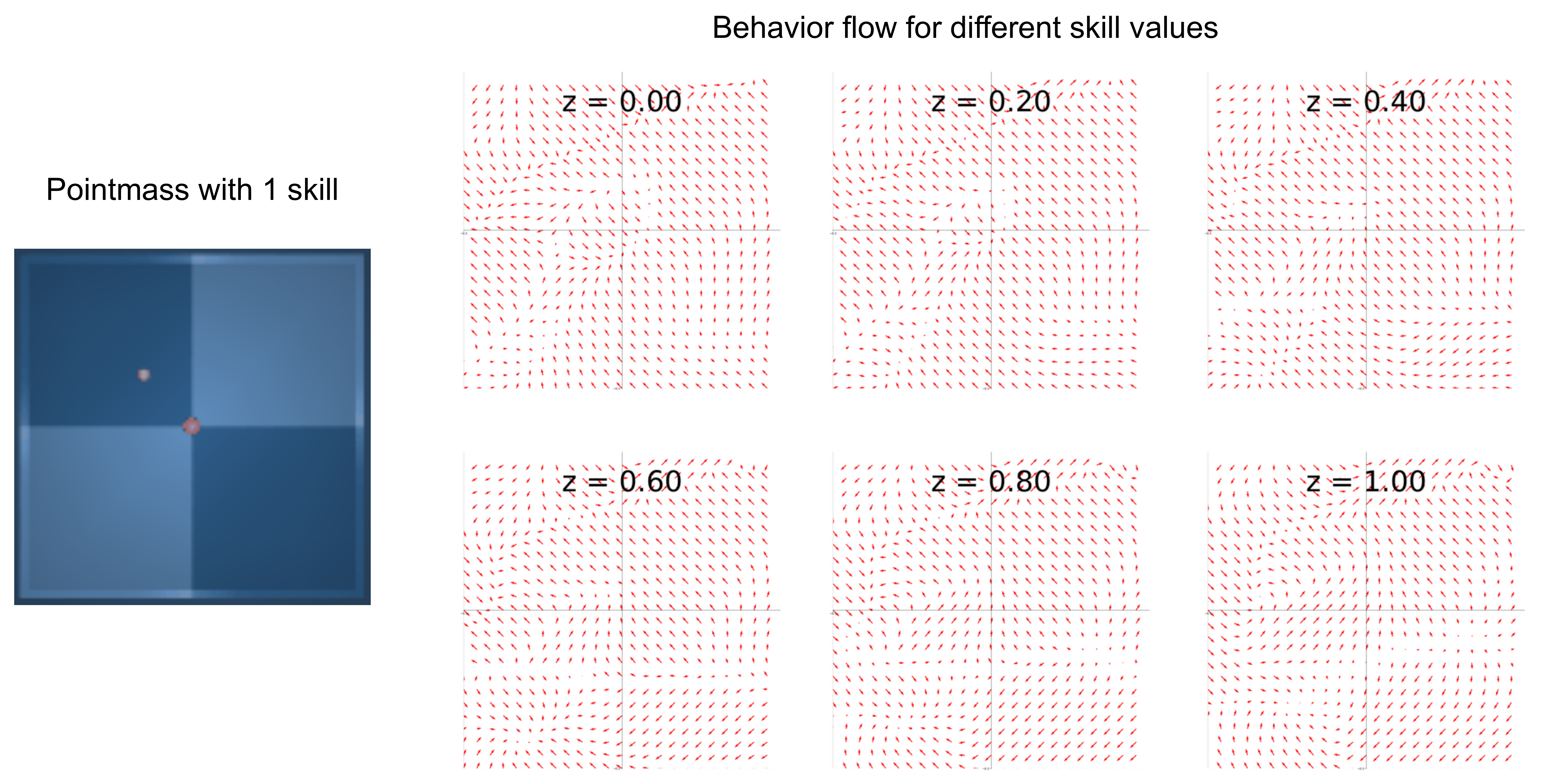}
\caption{\small Learning curves for finetuning pre-trained agents for 100k steps. Task performance is aggregated for each domain, such that each curve represents the mean normalized scores over $4 \times 10 = 40$ seeds. The shaded regions represent the standard error. \name surpasses the performance of the prior state-of-the-art on Walker and Jaco tasks while tying on Quadruped. \name is the only algorithm that performs consistently well across all three domains.
}
\label{fig:pmflow}
\end{figure*}

We provide two additional qualitative analyses of behaviors learned with the \name algorithm. First, we take a simple pointmass setting and set the skill dimension to 1 in order to ablate the skills learned by the \name agent in a simple setting. We sweep over different values of $z$ and plot the behavioral flow vector field (direction in which point mass moves) in Fig.\ref{fig:pmflow}. We find that the pointmass learns skills that produce continuous motion and that the direction of the motion changes as a function of the skill value. Near the origin the pointmass learns skills that span all directions, while near the edges the point mass learns to avoid wall collisions. Qualitatively, many behaviors are periodic.

\begin{figure*} [h] \centering
\includegraphics[width=.8\textwidth]{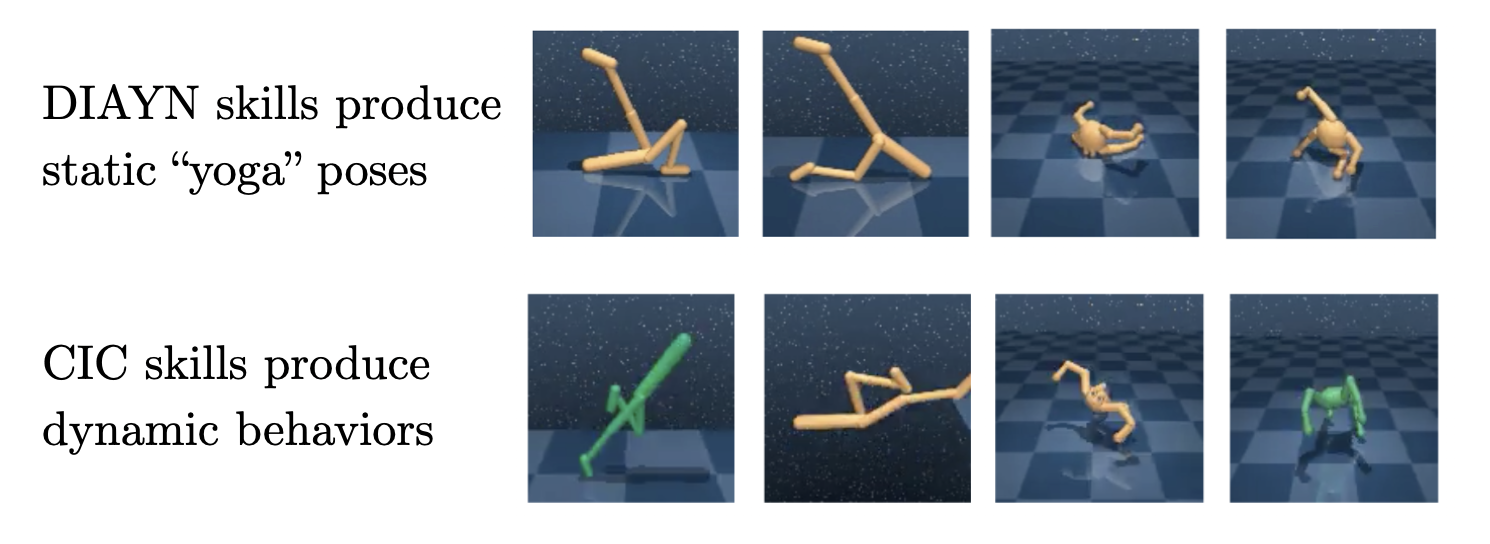}
\caption{\small Qualitative visualization of DIAYN and \name pre-training on the Walker and Quadruped domains from URLB. Confirming findings in prior work~\cite{zahavy2021}, we also find that DIAYN policies produce static but non-trivial behaviors mapping to ``yoga" poses while \name produces diverse and dynamic behaviors such as walking, flipping, and standing. Though it's hard to see from these images, all the DIAYN skills get stuck in frozen poses while the CIC skills are producing dynamic behavior with constant motion.
}
\label{fig:static}
\end{figure*}

Qualitatively, we find that methods like DIAYN that only support low dimensional skill vectors and do not explicitly incentivize diverse behaviors in their objective  produce policies that map skills to a small set of static behaviors. These behaviors shown in Fig.~\ref{fig:static} are non-trivial but also have low behavioral diversity and are not particularly useful for solving the downstream task. This observation is consistent with~\citet{zahavy2021} where the authors found that DIAYN maps to static ``yoga" poses in DeepMind Control. In contrast, behaviors produce by \name are dynamic resulting flipping, jumping, and locomotive behaviors that can then be adapted to efficiently solve downstream tasks.

\section{OpenAI Gym vs. DeepMind control: How Early Termination Leaks Extrinsic Signal}
\label{app:gymvsdmc}

Prior work on unsupervised skill discovery for continuous control~\citep{EysenbachGIL19diayn,SharmaGLKH20dads} was evaluated on OpenAI Gym~\citep{brockman2016openai} and showed diverse exploration on Gym environments. However, Gym environment episodes terminate early when the agent loses balance, thereby leaking information about the extrinsic task (e.g. balancing or moving). However, DeepMind Control (DMC) episodes have a fixed length of 1k steps. In DMC, exploration is therefore harder since the agent needs to learn to balance without any extrinsic signal. 

To evaluate whether the difference in the two environments has impact on competence-based exploration, we run DIAYN on the hopper environments from both Gym and DMC. We compare to ICM, a popular exploration baseline, and a Fixed baseline where the agent receives an intrinsic reward of 1 for each timestep and no algorithms receive extrinsic rewards. We then measure the extrinsic reward, which loosely corresponds to the diversity of behaviors learned. Our results in Fig.~\ref{fig:gymvsdmcskill} show that indeed DIAYN is able to learn diverse behaviors in Gym but not in DMC while ICM is able to learn diverse behaviors in both environments. Interestingly, the Fixed baseline achieves the highest reward on the Gym environment by learning to stand and balance. These results further motivate us to evaluate on URLB which is built on top of DMC.  

\section{CIC vs Other Types of Contrastive Learning for RL}

 Contrastive learning in CIC is different than prior vision-based contrastive learning in RL such as CURL~\cite{laskin2020curl}, since we are not performing contrastive learning over augmented images but rather over state transitions and skills. The contrastive objective in CIC is used for unsupervised learning of behaviors while in CURL it is used for unsupervised learning of visual features.

We provide pseudocode for the CIC loss below:
\begin{lstlisting}[language=Python, caption=CIC discriminator loss]
def discriminator_loss(states, next_states, skills, temp):
    """
    - states and skills are sampled from replay buffer
    - skills were sampled from uniform dist [0,1] during agent rollout
    - states / next_states: dim (B, D_state)
    - skills: dim (B, D_skill)
    """
    
    transitions = concat(states, next_states, dim=1)
    
    query = skill_net(skills) # (B, D_hidden) -> (B, D_hidden)
    key = transition_net(transitions) # (B, 2*D_state) -> (B, D_hidden)
    
    query = normalize(query, dim=1)
    key = normalize(key, dim=1)
    
    logits = matmul(query, key.T) / temp # (B, B)
    labels = arange(logits.shape[0])

    # positives are on diagonal, negatives are off diagonal 
    # for each skill, negatives are sampled from transitions 
    # while skills are fixed
    loss = cross_entropy(logits, labels)
    
    return loss
    
    
    


\end{lstlisting}

This is substantially different from prior contrastive learning works in RL such as CURL~\citep{laskin2020curl}, which perform contrastive learning over images.

\begin{lstlisting}[language=Python, caption=CURL contrastive loss]
def curl_loss(obs, W, temp):
    """
    - observation images are sampled from replay buffer
    - obs: dim (B, C, H, W)
    - W: projection matrix (D_hidden, D_hidden)
    """
    
    query = aug(obs)
    key = aug(obs)
    
    query = cnn_net(query) # (B, D_hidden)
    key = cnn_net(key) # (B, D_hidden)
    
    logits = matmul(matmul(query, W), key.T) / temp # (B, B)
    labels = arange(logits.shape[0])

    # positives are on diagonal
    # negatives are off diagonal
    loss = cross_entropy(logits, labels)
    
    return loss
    
    
    


\end{lstlisting}

\section{On estimates of Mutual Information}
\label{app:tightmi}

In this work we have presented \name - a new competence-based algorithm that achieves leading performance on URLB compared to prior unsupervised RL methods. 

 One might wonder whether estimating the exact mutual information (MI) or maximizing the tightest lower bound thereof is really the goal for unsupervised RL. In unsupervised representation learning, state-of-the-art methods like CPC and SimCLR maximize the lower bound of MI based on Noise Contrastive Estimation (NCE). However, as proven in CPC~\citep{oord2018representation} and illustrated in~\citet{poole2019mi} NCE is upper bounded by $\log N$, meaning that the bound is loose when the MI is larger than $\log N$. Nevertheless, these methods have been repeatedly shown to excel in practice. In ~\citet{tschannenDRGL20} the authors show that the effectiveness of NCE results from the inductive bias in both the choice of feature extractor architectures and the parameterization of the employed MI estimators.

We have a similar belief for unsupervised RL - that with the right parameterization and inductive bias, the MI objective will facilitate behavior learning in unsupervised RL. This is why CIC lower bounds MI with (i) the particle based entropy estimator to ensure explicit exploration and (ii) a contrastive conditional entropy estimator to leverage the power of contrastive learning to discriminate skills. As demonstrated in our experiments, CIC outperforms prior methods, showing the effectiveness of optimizing an intrinsic reward with the CIC MI estimator.

\section{Limitations}
\label{app:limitations}

While \name achieves leading results on URLB, we would also like to address its limitations. First, in this paper we only consider MDPs (and not partially observed MDPs) where the full state is observable. We focus on MDPs because generating diverse behaviors in environments with large state spaces has been the primary bottleneck for competence-based exploration. Combining \name with visual representation learning to scale this method to pixel-based inputs is a promising future direction for research not considered in this work. Another limitation is that our adaptation strategy to downstream tasks requires finetuning. Since we learn skills, it would be interesting to investigate alternate ways of adapting that would enable zero-shot generalization such as learning generalized reward functions during pre-training.

\end{document}